\documentclass[journal,hideappendix]{vgtc}
\usepackage{graphicx} % Required for inserting images
\usepackage[style=numeric,sorting=none,backend=biber]{biblatex}
\usepackage[utf8]{inputenc}
\usepackage{amsmath}
\usepackage{booktabs}
\usepackage{pifont}
\usepackage{tikz}
\usepackage{forest}
\usepackage{amssymb}
\usepackage{listings}
\usepackage{authblk}
\usepackage{multirow}
\usepackage{array} 

\usetikzlibrary{trees,positioning,shapes,shadows,arrows.meta}

\newcommand{\cmark}{\ding{51}} 
\newcommand{\xmark}{\ding{55}}

\newenvironment{j_itemize}{
\begin{itemize}[]
  \setlength{\itemsep}{3pt}
  \setlength{\parskip}{0pt}
  \setlength{\parsep}{0pt}
}{\end{itemize}}

\title{Human-inspired Perspectives: A Survey on AI Long-term Memory}

\author{
\textbf{Zihong He}\textsuperscript{1} \thanks{zihong.he@xintelligencelabs.ac}
\hspace{0.3cm} 
\textbf{Weizhe Lin}\textsuperscript{1,2} \thanks{weizhe.lin@xintelligencelabs.ac, Corresponding Author}
\hspace{0.3cm} 
\textbf{Hao Zheng}\textsuperscript{1}
\hspace{0.3cm}
\textbf{Fan Zhang}\textsuperscript{3} 
\hspace{0.3cm} 
\textbf{Matt W. Jones}\textsuperscript{3} 
\hspace{10cm}
\textbf{Laurence Aitchison}\textsuperscript{3}
\hspace{0.3cm}
\textbf{Xuhai Xu}\textsuperscript{4}
\hspace{0.3cm}
\textbf{Miao Liu}\textsuperscript{5}
\hspace{0.3cm}
\textbf{Per Ola Kristensson}\textsuperscript{2}
\hspace{0.3cm}
\textbf{Junxiao Shen}\textsuperscript{1,3}\thanks{junxiao.shen@bristol.ac.uk}
    \textsuperscript{1}~X-Intelligence Labs, 
    \textsuperscript{2}~University of Cambridge, 
    \textsuperscript{3}~University of Bristol,
    \textsuperscript{4}~Columbia University,
    \textsuperscript{5}~Meta
}

\abstract{
With the rapid advancement of AI systems, their abilities to store, retrieve, and utilize information over the long term - referred to as long-term memory - have become increasingly significant. These capabilities are crucial for enhancing the performance of AI systems across a wide range of tasks. However, there is currently no comprehensive survey that systematically investigates AI's long-term memory capabilities, formulates a theoretical framework, and inspires the development of next-generation AI long-term memory systems. 
This paper begins by introducing the mechanisms of human long-term memory, then explores AI long-term memory mechanisms, establishing a mapping between the two. Based on the mapping relationships identified, we extend the current cognitive architectures and propose the Cognitive Architecture of \textbf{S}elf-\textbf{A}daptive \textbf{L}ong-term \textbf{M}emory (\textbf{SALM}). \textbf{SALM} provides a theoretical framework for the practice of AI long-term memory and holds potential for guiding the creation of next-generation long-term memory driven AI systems. Finally, we delve into the future directions and application prospects of AI long-term memory.
}

\teaser{
  \centering
  \includegraphics[scale=0.5]{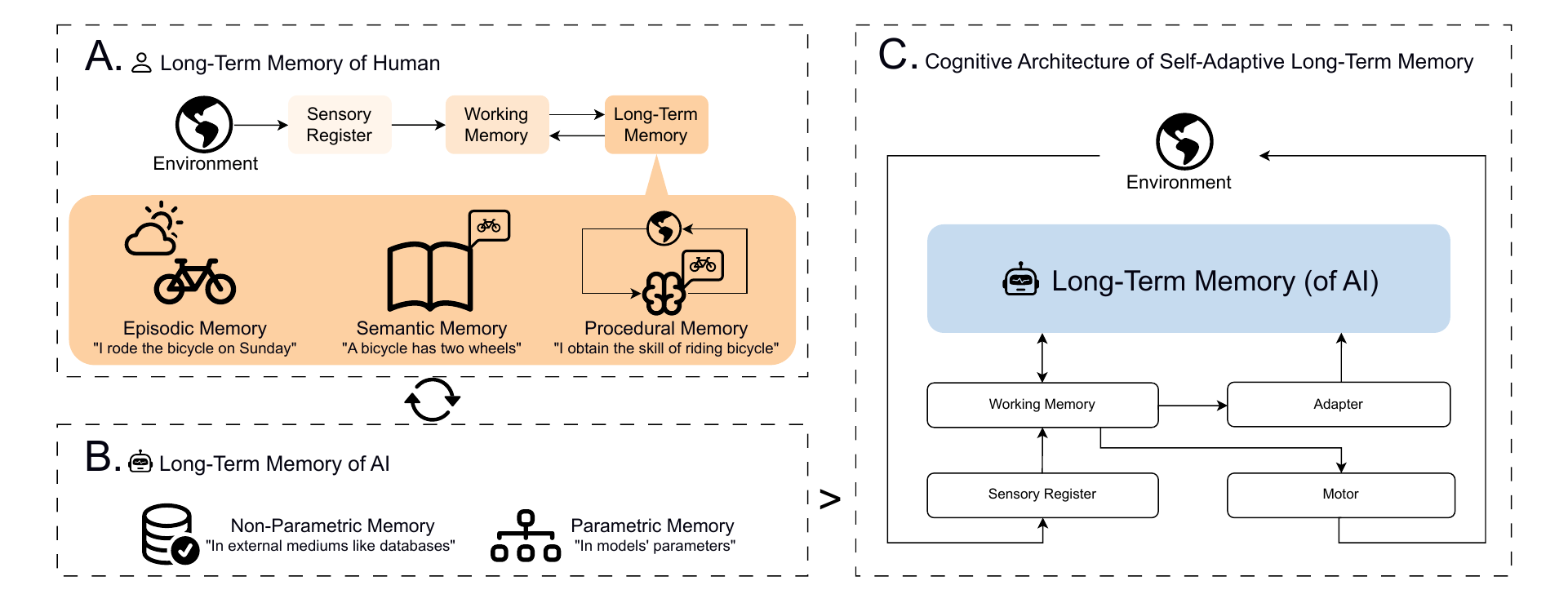}
  \caption{The mapping relationship between long-term memory of human (Part A in Fig.~\ref{fig:concepts} and discussed in Sec.~\ref{sec:human_memory}) and long-term memory of AI (Part B in Fig.~\ref{fig:concepts} and discussed in Sec.~\ref{sec:founddation_ai_memory} and Sec.~\ref{sec:hierarchy_ai_memory}); And the Cognitive Architecture of \textbf{S}elf-\textbf{A}daptive \textbf{L}ong-term \textbf{M}emory (\textbf{SALM}) proposed by us is based on theories related to long-term memory of AI (Part C in Fig.~\ref{fig:concepts} and discussed in Sec.~\ref{sec:hierarchy_ai_memory:cognitive_arc}).}
  \label{fig:concepts}
}

\begin{document}

\maketitle

\tableofcontents

\newpage
\section{Introduction}

\setcounter{tocdepth}{2}

\vspace{2\baselineskip} 
\begin{center}
    \textit{\textbf{``Memory is always reconstructed.''}}
    \begin{flushright}
    --- Geoffrey Hinton
    \end{flushright}
\end{center}

\begin{figure*}
\centering
\includegraphics[scale=0.4]{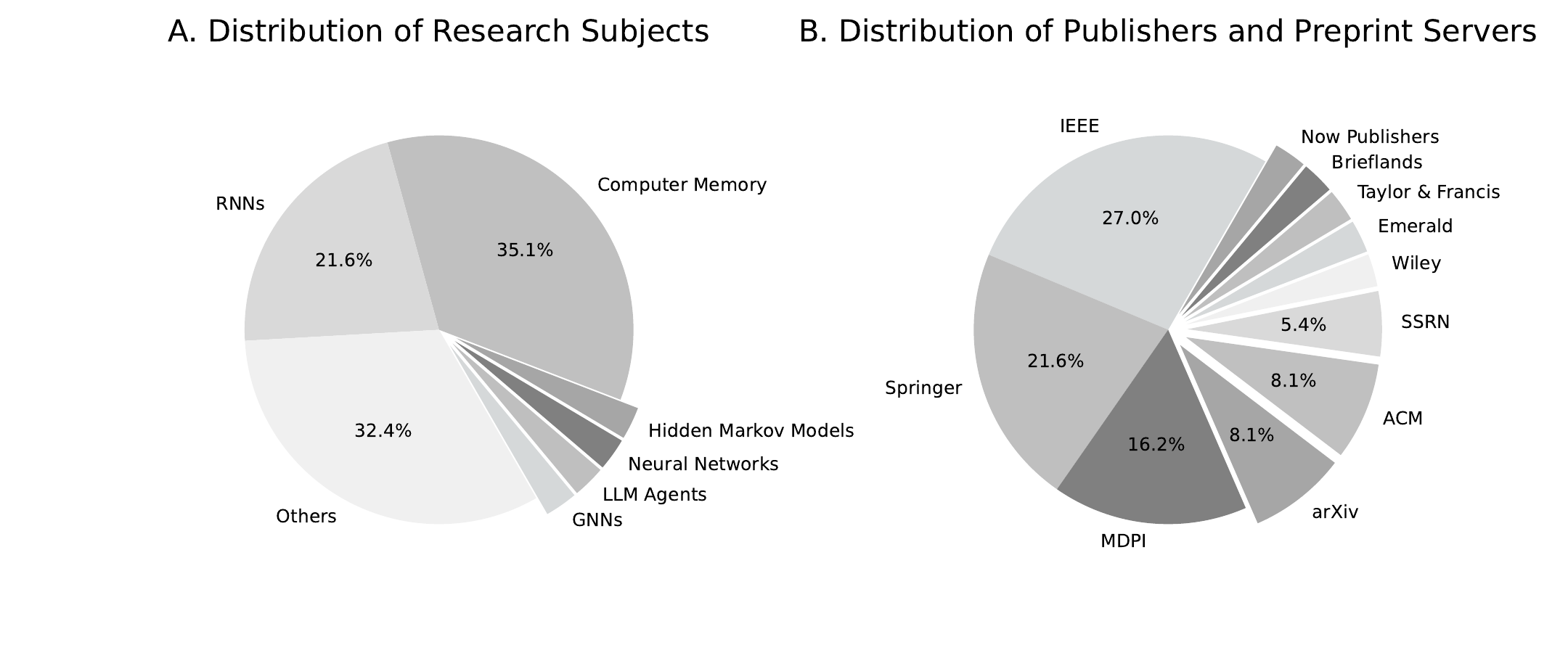}
\caption{Distributions of the research subjects, publishers and preprint servers of 37 papers related to AI memory Survey.}
\label{fig:research_distribution}
\end{figure*}

Over the past few decades, AI has demonstrated significant development potential and astonishing capabilities, thanks to the computational power supported by hardware and advanced machine learning algorithms~\cite{hochreiter1997long,nilsson2009quest,vaswani2017attention,xi2023agents,wayne2023llm}. As AI evolves and its applications span various industries, endowing it with long-term memory becomes increasingly important. Long-term memory in AI systems can be understood by analogy with human long-term memory, which stores information over periods ranging from days to years~\cite{atkinson1968human}. Long-term memory assists AI systems in several ways. First, it helps acquire general understanding, such as integrating learned knowledge to enhance question-answering systems~\cite{qa2020karpukhin,xixin2022gmtkbqa,yang2023care,luo2023chatkbqa}. Additionally, it establishes connections between current and past scenarios by storing key features from frames for timely retrieval, benefiting applications like video understanding~\cite{cheng2022xmem,mezghan2022memory,shen2023encode,song2023moviechat,he2024ma}. Furthermore, it aids in mastering procedural abilities, such as developing strategy selection capabilities through production rules or reinforcement learning, thus enhancing agents' adaptability~\cite{davis1977production,quinlan1987generating,ceri1991deriving,kaelbling1996reinforcement,laird2008extending,liu2021automated,liu2022feature}.

The human memory system offers valuable insights for designing AI long-term memory, as evidenced by recent AI developments that have, to varying degrees, emulated and incorporated structures akin to those in human long-term memory. For instance, some video understanding systems~\cite{cheng2022xmem,song2023moviechat} draw upon the Atkinson-Shiffrin Model~\cite{atkinson1968human} (a highly influential human memory model, detailed in Sec.~\ref{sec:human_memory}) to construct hierarchical memory systems. 
These systems integrate short-term memory, which temporarily stores visual information, with long-term memory, which recalls historical data to enhance task effectiveness. 
Moreover, cognitive architectures introduced in works such as ~\cite{laird2008extending,sumers2023cognitive} adopted pivotal components of human long-term memory: episodic, semantic, and procedural memory (also discussed in Sec.~\ref{sec:human_memory}). These cognitive architectures leverage the history event flow to form episodic memory, knowledge sources to form semantic memory, and production rules, code or reinforcement learning to form procedural memory, enabling the creation of agents capable of self-learning and experience accumulation.
Moreover, other works, although not explicitly grounded in human long-term memory theories, employ methodologies that closely mirror the processing mechanisms of human long-term memory, as explored later in Sec.~\ref{sec:founddation_ai_memory} and Sec.~\ref{sec:hierarchy_ai_memory}. 
Therefore, the principles underlying human long-term memory present a promising theoretical foundation for AI researchers to develop more advanced AI systems with robust and flexible long-term memory capabilities.

Our research identifies a gap in the existing literature, as there are no comprehensive surveys on AI long-term memory that are grounded in human memory theories (Sec.~\ref{sec:survey_ltm_ai}).
To bridge this gap, this paper provides a survey of AI long-term memory from the perspective of human memory theory, to motivate and constrain prototypes, taxonomies, associated challenges, and system design of AI long-term memory. 
This survey offers researchers with valuable insights for constructing and optimizing AI systems that utilize long-term memory.
The main content of this survey is outlined as follows.

In Sec.~\ref{sec:human_memory}, we introduce human memory categorized by stages of information processing~\cite{atkinson1968human}, focusing on episodic memory, semantic memory, and procedural memory, the three key subsystems of human long-term memory~\cite{tulving1972episodic,tulving1973encoding,tulving1985memory}. 
We then classify AI long-term memory into parametric and non-parametric memory in Sec.~\ref{sec:founddation_ai_memory}, based on the form of storage~\cite{lewis2020retrieval,mallen2023trust}. 
% Non-parametric memory is explicitly stored in external media outside the AI model, while parametric memory is implicitly kept within the model's parameters. 
We also discuss the processing mechanisms and the challenges associated with these different formats. In Sec.~\ref{sec:hierarchy_ai_memory}, we make an attempt to establish the relationships between AI's and human's long-term memory based on the review. 
Based on these relationships, in Sec.~\ref{sec:hierarchy_ai_memory:cognitive_arc}, we proposed the Cognitive Architecture of \textbf{S}elf-\textbf{A}daptive \textbf{L}ong-term \textbf{M}emory (\textbf{SALM}) by integrating theories of AI long-term memory.
This framework addresses the limitations of long-term memory modules in current cognitive architectures and has the potential to exceed the adaptability of human long-term memory processing mechanisms~\cite{shiffrin1969memory,sweller2003evolution,manns2006evolution,allen2013evolution}. It has the promise of serving as a guiding framework for the next generation of AI systems driven by long-term memory. 
Finally, in Sec.~\ref{sec:next_steps}, we introduce practical measurement methods and potential applications for AI long-term memory.

In summary, the key contributions of our survey are:
\begin{j_itemize}
\item[$\bullet$] We conduct a narrative review of the convergent work on long-term memory of both human and AI.
\item[$\bullet$] We develop a taxonomy of AI long-term memory based on the human memory theories, highlighting the strong associative relationships between these two.
\item[$\bullet$] We propose a cognitive architecture that integrates AI long-term memory theories with adaptive mechanisms for long-term memory processing, offering a potential guiding framework for the next generation of AI systems driven by long-term memory.
\item[$\bullet$] We examine the metrics and applications for AI long-term memory modules, advancing the implementation of AI systems driven by long-term memory.
\end{j_itemize}

\section{Research Background and Methodologies}
\label{sec:survey_ltm_ai}

\textbf{Research Background.} We conduct a comprehensive analysis of articles from 2015 onward, relevant to AI memory surveys. Our search, conducted on October 7th, 2024, utilized the following search terms: \textit{(Review(s) or Survey(s) or Taxonomy(ies)) and Memory(ies) and (AI or Artificial Intelligence or Agent(s) or Deep Learning or Machine Learning or Neural Network(s))}, and organized the research subjects, related publishers and preprint servers of these articles, as depicted in Fig.~\ref{fig:research_distribution}. The majority of the reviewed articles focus on computer memory and recurrent neural networks (RNNs). The computer memory domain primarily addresses data storage and retrieval, rather than exploring memory taxonomy from an AI perspective~\cite{mittal2018survey,umesh2019survey,vurdelja2020survey,radanliev2021review,qian2021graph,asad2022survey,chen2022resistive,haensch2023compute,snasel2023review,ulidowski2023saving,radhika2023review,kaur2024comprehensive,alaparthi2024survey}. In contrast, RNNs, particularly long short-term memory networks (LSTMs), focus on managing hidden states for storing and utilizing sequential historical information, which is not generalizable across all AI domains~\cite{bagherzadeh2019review,ma2019taxonomy,li2021quantitative,su2022recurrent,tan2022performance,khan2023short,ghojogh2023recurrent,tyagi2023comparative}. The survey by Savya et al.~\cite{savya2023memory} provides a broader analysis of neural network models, including Transformers~\cite{vaswani2017attention} and Neural Turing Machines~\cite{graves2014neural}, and examines the implications of human memory theories, such as the Atkinson-Shiffrin Model~\cite{atkinson1968human}, for AI memory. However, they focused on the implementation of memory within specific model architectures, and they did not further categorize long-term memory into types such as episodic, semantic, or procedural memory~\cite{tulving1985memory} to assess their relevance to AI. On the other hand, the development of Large Language Models (LLMs)~\cite{wayne2023llm} has recently accelerated research on memory in intelligent agents. Zhang et al.~\cite{zhang2024survey} categorized the memory types in LLM-based agents into textual and parametric forms, which aligned with the long-term memory characteristic of retaining information for an extended period~\cite{atkinson1968human}. However, their work does not explicitly incorporate human memory theories to structure the discussion. In summary, current surveys on memory in the field of AI, particularly long-term memory, face several limitations:
  \begin{j_itemize}
  \item[$\bullet$] Lack a focus on AI systems as a whole. 
  \item[$\bullet$] Lack a framework that incorporates theories of human memory.
  \end{j_itemize}

\noindent\textbf{Research Methodologies.} To address these limitations, we adopt human memory theories as the foundation for structuring our review of AI long-term memory. 
  Our approach begins with a review of literature on human memory, focusing on its stages of processing as discussed in cognitive science and neuroscience (Sec.~\ref{sec:human_memory}). We then examine the hierarchy and processing mechanisms related to human long-term memory (Fig.~\ref{fig:Hierarchy_human}), using these as prototypes to identify AI-related studies that exhibit comparable characteristics (Sec.~\ref{sec:founddation_ai_memory} and Sec.~\ref{sec:hierarchy_ai_memory}).
  Many AI papers related to long-term memory do not explicitly label ``long-term memory'' as a topic, instead incorporating it implicitly. For instance, numerous studies on long-term memory retrieval, as discussed in Sec.~\ref{sec:AI_memory:non_parametric:memory_retrieval}, may not explicitly reference ``memory'' in their themes; however, they remain intrinsically linked to the retrieval processes associated with human long-term memory. Therefore, in addition to the search for terms like ``memory'' or ``long-term memory'', our review also examines common AI themes such as ``neural networks'', ``deep learning'', ``reinforcement learning'', and more recent trends like ``LLMs'' and ``RAG'' (Retrieval-Augmented Generation), as well as interdisciplinary fields like ``cognitive architectures''. We classify and synthesize our findings, connecting them to the processing and classification systems of human long-term memory as outlined in Sec.~\ref{sec:founddation_ai_memory} and Sec.~\ref{sec:hierarchy_ai_memory}, and illustrated in Fig.~\ref{fig:lit_surv}. Additionally, we propose an actionable framework based on our human memory-inspired AI long-term memory theory (Sec.~\ref{sec:hierarchy_ai_memory:cognitive_arc}) and discussed relevant metrics and applications powered by AI long-term memory (Sec.~\ref{sec:next_steps}). We identify several representative papers to substantiate each point of discussion.

In summary, this survey was conducted to address the existing gap in summarizing and analyzing AI long-term memory through the lens of human memory theories. By comparing AI long-term memory with categories and processing methods of human long-term memory, we review relevant representative work, as shown in Fig.~\ref{fig:lit_surv}.

\section{Long-term Memory in Human Brain}
\label{sec:human_memory}

\begin{figure*}
\centering
\includegraphics[]{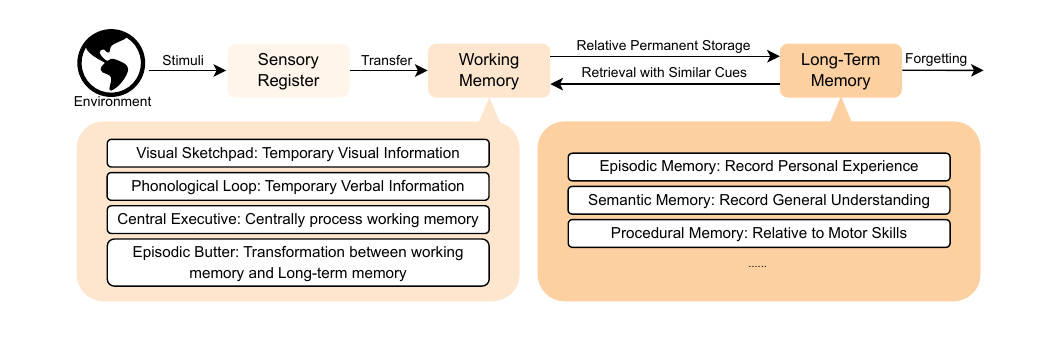}
\caption{Overview of Human Memory Hierarchy and processing. The \textbf{Sensory Register} receives and temporarily stores information; \textbf{Working Memory} stores critical information for immediate tasks; and \textbf{Long-term Memory} stores information relatively permanently.}
\label{fig:Hierarchy_human}
\end{figure*}

Human long-term memory is a subject of extensive research spanning various fields, including cognitive psychology, neuroscience, and computational neuroscience. Insights into human long-term memory enhance our understanding of brain function for researchers in these domains and also offer valuable guidance for AI researchers in constructing effective AI long-term memory. In this section, we provide an overview of human memory's hierarchy (Sec.~\ref{sec:human_memory:hierarchy}) and processing (Sec.~\ref{sec:human_memory:processing}) that are closely related to long-term memory. 
The overview of the human memory system is illustrated in Fig.~\ref{fig:Hierarchy_human}.

\subsection{Human Memory Hierarchy}
\label{sec:human_memory:hierarchy}
 
Before introducing the hierarchy of human memory, it is important to note the close connection between research on human memory and the development of cognitive psychology. 
Traditional Behaviorism focuses on the environment and observable behaviors~\cite{watson2017behaviorism}, while cognitive psychology, established in the 1960s, emphasizes the ``internal system'' involved in the formation of mental phenomena~\cite{neisser2014cognitive}. This shift is considered a major breakthrough in psychology and has spurred extensive research on human memory.
Several influential theories have emerged from the development of cognitive psychology, including the ``Levels of Processing'' theory, which highlights the relationship between memory processing and its effects~\cite{lockhart1990levels}, the ``Working Memory'' theory, which focuses on the active maintenance and manipulation of information~\cite{baddeley1974working}, and the ``Atkinson-Shiffrin model''~\cite{atkinson1968human}, which emphasizes the hierarchical structure of human memory.

Among these models, Atkinson-Shiffrin model, proposed by Atkinson et al., represents a significant milestone in memory research. It is recognized for its foundational and comprehensive approach to explain the hierarchical division of memory and its substantial influence on subsequent research in the field. 
Consequently, we use this model to exemplify the theoretical basis for explaining the hierarchy of human memory.
Notably, this model and its derivative theories are frequently referenced in AI memory research~\cite{lei2023recagent, cheng2022xmem, song2023moviechat}, bridging the fields of human and AI memory studies.
The Atkinson-Shiffrin model categorizes the human memory system into three levels: Sensory Register (Sec.~\ref{sec:human_memory:hierarchy:sensor_register}), Short-term Store (Working Memory)(Sec.~\ref{sec:human_memory:hierarchy:short_term_memory}), and Long-term Store (Long-term Memory) (Sec.~\ref{sec:human_memory:hierarchy:long_term_memory}).

\subsubsection{Sensory Register}
\label{sec:human_memory:hierarchy:sensor_register}

According to Atkinson et al.~\cite{atkinson1968human}, the sensory register is responsible for receiving and temporarily storing information from sensory systems, allowing information like visual stimuli to be retained in a highly precise form for a short period. They used Sperling et al.'s 1960 visual exposure experiment \cite{sperling1960information} to illustrate the existence of the sensory register. In Sperling et al.'s experiment, subjects were briefly exposed to a series of printed letters and then asked to recite them after exposure. They discovered that for exposures lasting between 15 to 500 milliseconds, subjects could correctly report an average of slightly over four letters. When the stimuli contained four or fewer letters, subjects could report them with almost 100\% accuracy.
This finding demonstrated that individuals can retain brief visual stimuli through a mechanism that reflects the function of the sensory register.
% This information is likely encoded in firing of neurons, or potentiation of synases in low-level sensory areas such as V1.
Information stored in the sensory register rapidly decays and either vanishes or is transferred to short-term memory within a very short time, providing raw material for subsequent memory processing~\cite{atkinson1968human}.
This underscores the sensory register's critical role in the initial stages of information processing and memory formation.

\subsubsection{Working Memory}
\label{sec:human_memory:hierarchy:short_term_memory}

In 1968, Atkinson et al.~\cite{atkinson1968human} summarized the concept of ``short-term storage'' based on previous research findings. These findings included the work of Peterson et al.~\cite{peterson1959short}, who observed variations in participants' ability to recall consonant letters\footnote{Consonant letters are letters of the alphabet that are not vowels (a, e, i, o, u).} at different time intervals, and Milner et al.~\cite{milner1959memory, milner1966amnesia, milner1966neuropsychological}, who discovered that patients who had their bilateral hippocampal region removed could recall short-term information like normal individuals, but could not retain new long-term information. Atkinson et al. indicated that short-term storage temporarily holds information, and if this information is not maintained through control processes such as rehearsal, it will decay and vanish within a short period, approximately 30 seconds, as noted by Milner et al.~\cite{milner1966neuropsychological}.

However, the short-term storage theory proposed by Atkinson et al. has limitations in explaining complex human information processing, particularly regarding how information is processed and stored within the short-term store. They did not systematically introduce the division and details of information processing in the short-term store. Addressing these limitations, Baddeley et al.~\cite{baddeley1974working} proposed the working memory theory in 1974, suggesting that working memory is used for the temporary storage and manipulation of information. For example, when performing mental arithmetic, one must not only store numbers temporarily but also apply arithmetic operations to them sequentially. 
Another example is that recalling numerical information from an oral narrative and completing a reasoning task both require storing and manipulating the information within working memory.
Similarly, taking language comprehension to exemplify working memory, individuals must hold words and phrases in memory while simultaneously integrating them into larger syntactic structures and deriving meaning.
The manipulations in these examples are essential for tasks like problem-solving, decision-making, and reasoning, which rely on the dynamic handling of information within working memory~\cite{baddeley1974working}. From 1974 to 2000, a series of works by Baddeley et al.~\cite{baddeley1974working,baddeley1986working,baddeley2000episodic} expanded the working memory theory, dividing it into four components: the Central Executive, Phonological Loop, Visuospatial Sketchpad, and Episodic Buffer. 
We explain these components briefly while summarizing them in Fig.~\ref{fig:Hierarchy_human}:

\begin{j_itemize}

    \item[$\bullet$] \textbf{Central Executive.} Baddeley et al.~\cite{baddeley1974working,baddeley1986working} highlighted that the central executive is a critical component of working memory, designed to control and supervise the operation of various modules within working memory, playing a pivotal role in resource allocation similar to that of a CPU (Central Processing Unit) in a computer. When individuals need to process complex transient information, such as temporarily remembering a long sequence of phone numbers, the central executive directs the individual to concentrate more on this task.

    \item[$\bullet$] \textbf{Phonological Loop.} Baddeley et al.~\cite{baddeley1986working} further categorized the phonological loop as a type of working memory, aiming to represent humans' transient memory for verbal information. The phonemic loop refers to the looping of phonemes in the brain, like an ``inner ear'' that holds information in a speech-based form for a short time. It is responsible for storing words we hear and silently repeat.
    The phonemic buffer temporarily stores phonological information, such as phrases, sentences, and numbers. This buffer acts like an ``inner voice'', enabling us to repeat information in a loop to maintain and refresh it, preventing it from fading. A demonstration of the phonological loop in action is the process of memorizing a string of phone number by reciting it repeatedly (and silently)  until we are able to jot it down in a notebook.
    The phonological loop plays a crucial role in memorizing linguistic information, such as new words~\cite{baddeley1998phonological}.

    \item[$\bullet$] \textbf{Visuospatial Sketchpad.} According to Baddeley et al.~\cite{baddeley1986working}, the visuospatial sketchpad temporarily stores and processes visual and spatial information. It integrates visual and spatial information from perception with information stored in long-term memory. This helps in recognizing objects, understanding scenes, and performing tasks that require spatial awareness, such as driving, reading maps, engaging in sports, and visualizing and manipulating objects in mind. In 1994, the theory proposed by Kosslyn et al.~\cite{kosslyn1994image} provided an enhanced understanding of visuospatial sketchpad. This theory delineates that visuospatial sketchpad can be divided it into three steps: image generation, image maintenance, and image rotation. This process is analogous to solving geometry problems that involve rotating shapes. First, an individual creates an initial mental image of the shape. Next, they maintain the image in their mind. Finally, they imagine how the shape would rotate to match the target image. These researches highlight the critical role of the visuospatial sketchpad, demonstrating its importance in cognitive tasks that require mental manipulation of visual information.

    \item[$\bullet$] \textbf{Episodic Buffer.} Baddeley et al.~\cite{baddeley2000episodic} introduced the episodic buffer as a component of working memory. Episodic buffer facilitates the transfer of information between working memory and long-term memory. Specifically, the episodic buffer enables the movement of significant information from working memory to long-term memory and allows fragments of long-term memory that are relevant to the current context to be retrieved and placed into the episodic buffer. This process supports the formation of new long-term memory and the effective use of long-term memory.
    
\end{j_itemize}

\subsubsection{Long-term Memory}
\label{sec:human_memory:hierarchy:long_term_memory}

In 1968, Atkinson et al.~\cite{atkinson1968human} suggested that the short-term store includes a mechanism for transferring information to the long-term store. Unlike the sensory register and short-term store, information in the long-term store does not rapidly decay or vanish. This information is relatively permanent, although it can be modified by other information or become temporarily inaccessible.
According to Atkinson et al., the information in long-term store, referred to as ``traces'', does not strictly conform to the ``all-or-none'' characteristic. Instead, ``traces'' can exist in intermediate states such as ``decay'' and ``interference''. ``Decay'' refers to the weakening of long-term store ``traces'', such as information that deteriorates over time due to a lack of review, for example, certain words in a long sentence. ``Interference'' describes the mutual influence between ``traces'', such as when learning word A followed by a similar word B leads to confusion during recall. The term "long-term store" used by Atkinson et al. refers specifically to memory as a physical or functional analogy for information storage. In a broader context, this concept is referred to as ``long-term memory''.

In 1972, Tulving et al.~\cite{tulving1972episodic} highlighted the complex nature of long-term memory by introducing two distinct types: episodic memory and semantic memory. Episodic memory relates to personal past events and experiences, such as an experience of mountain climbing or a recent vacation. In contrast, semantic memory relates to general information, such as ``Paris is the capital of France'' or understanding the concept of gravity.
Furthermore, in 1985, Tulving et al.~\cite{tulving1985memory} classified memory involving the acquisition of perceptual feedback and motor skills as procedural memory~\cite{winograd1972understanding,tulving1983elements,anderson2013language}. An example of this is learning to ride a bicycle, which requires acquiring motor skills through balance feedback. Based on these existing findings, Tulving et al.~\cite{tulving1985memory} assumed three memory systems: procedural memory, semantic memory, and episodic memory.

Graf et al.~\cite{graf1985implicit} introduced an alternative method of memory classification based on whether the memory can be consciously recalled, categorizing memory into explicit and implicit types. Implicit memory operates without conscious recollection; for example, maintaining balance on a bicycle after learning to ride involves implicit memory, as it does not require active thought. In contrast, explicit memory necessitates conscious recollection, such as when recalling a specific experience or piece of knowledge while writing an article. 
According to the classification proposed by Tulving et al.~\cite{tulving1985memory}, episodic memory and semantic memory are types of explicit memory~\cite{tulving1972episodic}, whereas procedural memory falls under implicit memory~\cite{schacter1987implicit}. 

Having established the categorization of long-term memory types, we now discuss the specifics of episodic, semantic, and procedural memory and show them briefly in Fig.~\ref{fig:Hierarchy_human}:

\begin{j_itemize}

    \item[$\bullet$] \textbf{Episodic Memory.} Tulving et al.~\cite{tulving1972episodic} defined episodic memory in 1972 as a type of memory specific to individuals, related to time and space, involving personal experiences, events, and contexts. In 1973, Tulving et al.~\cite{tulving1973encoding} noted that input stimuli such as sound and visual information are processed for episodic memory storage through \textit{specific encoding}, which combines perceived information with specific contexts and experiences to form memory trace (discussed in the leading paragraph of Sec.~\ref{sec:human_memory:hierarchy:long_term_memory}) that aids in subsequent retrieval. For successful recall of this trace, similar input stimuli need to be provided. For example, adults often recall childhood experiences upon seeing childhood photos, which stimulates the recall of relevant long-term memory traces. 
    % In 1985, Tulving et al.~\cite{tulving1985memory} explored the relationship between memory and consciousness, suggesting that episodic memory is associated with \textit{autonoetic consciousness}. This form of consciousness allows individuals to associate the present situation with specific personal events from the past, with an awareness that the situation are genuine parts of their past experiences. 
    Furthermore, episodic memory also facilitates learning from past experiences and applying this knowledge to future decisions, thereby enhancing an individual's ability to adapt to their environment~\cite{tulving2002episodic}. The hippocampus in the brain plays a vital role in learning episodic memory by encoding them through a process that separates similar personal experiences and stores them distinctly, thereby minimizing interference between different episodic memory~\cite{marr1991simple,o1994hippocampal,o2014complementary}.

    \item[$\bullet$] \textbf{Semantic Memory.} According to Tulving et al.'s perspective~\cite{tulving1972episodic} in 1972, semantic memory encompasses an individual's knowledge about words and other linguistic symbols, including the relationships between them, and the rules, formulas, and algorithms for manipulating them. It functions to receive, retain, and convey semantic information about words and concepts. Unlike episodic memory, which is tied to personal experiences, semantic memory deals with general information. An example of semantic memory is the long-term memory formed by learning knowledge in a specific field or by extracting general rules from past experiences. 
    % In 1985, Tulving et al.~\cite{tulving1985memory} further explored the relationship between memory and consciousness, indicating that the type of consciousness associated with semantic memory is termed \textit{noetic consciousness}. Noetic consciousness is an internal, non-physical state of awareness that allows individuals to perceive certain objects without external stimuli and is more closely associated with intuition, insight, and internal knowledge.
    % For example, when a person recalls the definition of a word or a mathematical formula, they do so through noetic consciousness, which does not need to be called up based on a certain scene. This contrasts with \textit{autonoetic consciousness}, which connects the present scene to episodic memory. 
    The management of semantic memory involves multiple brain regions, such as the left prefrontal cortex (LPC) being involved in the retrieval of semantic information, and the temporal lobes storing specific information related to objects~\cite{hodges1992semantic,baldo1998letter,martin2001semantic}.

    \item[$\bullet$] \textbf{Procedural Memory.} Tulving et al.~\cite{tulving1985memory} describe procedural memory as the memory developed from learning motor skills through feedback and perceptual abilities. This type of memory is intrinsically linked to the immediate temporal and spatial contexts and does not depend on information beyond the current stimuli. This allows for instantaneous reactions based on direct perception. For example, learning to ride a bicycle involves the sensation of losing balance, which triggers discomfort and gradually leads to the development of balancing skills. Once these skills are fully acquired, individuals can effortlessly maintain balance while riding, demonstrating the role of procedural memory in learning motor skills. Brain regions such as the supplementary motor area (SMA), primary motor cortex (M1), prefrontal cortex, and cerebellum play crucial roles in skill consolidation for the formation of procedural memory~\cite{censor2014cortico}.

\end{j_itemize}

\subsection{Human Memory Processing}
\label{sec:human_memory:processing}

In the previous section, we divide human memory into sensory register, short-term store, and long-term store (long-term memory) following the Atkinson-Shiffrin Model and its related theories. These theories explain human memory in terms of its hierarchical structure. 
However, understanding human memory can also be approached from the perspective of the entire memory processing cycle. This perspective allows for a deeper investigation into the ``black box'' of the brain by dividing the entire process into separate stages.

Since the mid-19th century, advancements in brain science, computational neuroscience, and cognitive psychology have given rise to theories related to memory processing. 
In 1963, Melton et al.~\cite{melton1963implications} emphasized the significance of \textbf{memory storage} and \textbf{memory retrieval} in memory theory research. Storage refers to the mechanism by which memory is placed in the brain, while retrieval refers to the mechanism by which memory is recalled from the brain. The units of memory storage, known as ``traces'', can be strengthened through active rehearsal or consistent use~\cite{muller1900experimentelle}. Conversely, unused memory traces will decay over time~\cite{thorndike1913educational}. Melton et al. also suggested that the storage of memory traces is closely related to their subsequent retrieval, which may be influenced by factors such as trace integrity, interference, and repetition. 
Interference can compromise the integrity of memory traces, with lower integrity leading to less effective retrieval. For example, learning a new word may interfere with the memory of a similar old word, affecting its usage in writing. 
Repetition, on the other hand, can enhance the strength of memory traces, thereby improving retrieval efficiency; for instance, repeated review before an exam leads to better grades. Additionally, in 1969, Atkinson et al.~\cite{shiffrin1969memory} proposed that memory storage and retrieval in long-term memory function as parallel processes that correspond to each other. Later, in 1973, Tulving et al.~\cite{tulving1973encoding} noted that memory storage and retrieval are distinct research directions.
Based on these studies, we establish two independent subsections, Sec.~\ref{sec:human_memory:processing:storage} and Sec.~\ref{sec:human_memory:processing:retrieval}, to elaborate on the processes of storage and retrieval in human memory in detail.

Beyond storage and retrieval, forgetting also plays a vital role in the understanding of memory. Specifically, the research by Ebbinghaus et al.~\cite{ebbinghaus1885memory} in the late 19th century sparked continued interest among subsequent neuroscientists in the field of \textbf{memory forgetting}, leading to a series of experimental analyses and ongoing attention~\cite{jost1897law, wixted1997genuine, anderson1997memory}.

From a neuroscience perspective, experimental analyses of cellular mechanisms in memory processing continue to uncover the anatomical and physiological foundations of memory. Many studies highlight the critical roles of the hippocampus, the neocortex, and their interaction in supporting long-term memory~\cite{marr1991simple,lavenex2000hippocampal,martin2001semantic,ji2007coordinated,o2014complementary,censor2014cortico} (Discussed in Sec.~\ref{sec:human_memory:hierarchy:long_term_memory} and the subsequent subsections). In brains, memory stability necessarily reflects the lifecycle of neurons. For example, Deisseroth et al.~\cite{deisseroth2004excitation} demonstrated that excitatory stimuli can increase the proportion of neural stem cells/progenitor cells (NPCs) in the adult hippocampus differentiating into neurons. The genesis of new neurons may participate in the storage of new memory, and their addition may accompany the replacement of synaptic connections of old neurons, thereby accelerating the forgetting of old memory. Meanwhile, different subfields of the hippocampus have been implicated in supporting different memory modalities. For example, dorsal CA2 of the mouse hippocampus may preferentially contribute to the encoding and recall of social memories related to the identity of conspecifics~\cite{meira2018hippocampal}.  
Furthermore, Weinberger et al.~\cite{weinberger2004specific} demonstrated in 2004 that the primary auditory cortex can acquire and retain specific memory traces about the behavioral significance of certain sounds, the plasticity changes in the cortex strengthens over time and specific memory-related sound stimuli can activate particular neural networks within the primary auditory cortex. This highlights the cortex’s crucial roles in the storage, consolidation, and retrieval of long-term memory.

In summary, memory processing is a multifaceted research topic that spans multiple disciplines. It can generally be divided into three key processes: memory storage, memory retrieval, and memory forgetting. By understanding the mechanisms behind these processes, we can gain deeper insights into the functioning of the human brain.
In the following sections, we will provide detailed explanations for memory processing, focusing on its three parts: Memory Storage (Sec.~\ref{sec:human_memory:processing:storage}), Memory Retrieval (Sec.~\ref{sec:human_memory:processing:retrieval}), and Memory Forgetting (Sec.~\ref{sec:human_memory:processing:forgetting}).
The processing mechanisms and their interactions with other memory components are demonstrated in Fig.~\ref{fig:Hierarchy_human}.

\subsubsection{Memory Storage}
\label{sec:human_memory:processing:storage}

Based on the perspective of Tulving et al.~\cite{tulving1973encoding}, memory storage is achieved by processing input stimuli through \textit{specific encoding} (detailed in the episodic part of Sec.~\ref{sec:human_memory:hierarchy:long_term_memory}).
The hippocampus and cortex of the brain are involved in the encoding of memory, with the hippocampus encoding unique episodic information through its pattern separation properties and the cortex encoding more generalized information through its distributed representations~\cite{marr1991simple,o1994hippocampal,winocur2010memory,o2014complementary}.
Sensory register (Sec.~\ref{sec:human_memory:hierarchy:sensor_register}), short-term store (Sec.~\ref{sec:human_memory:hierarchy:short_term_memory}), and long-term store (Sec.~\ref{sec:human_memory:hierarchy:long_term_memory}) have distinct methods for storing and encoding information at different stages of the memory process~\cite{shiffrin1969memory}. 
The long-term store, in particular, encodes memory through four distinct encoding strategies: ``meaning'', ``association'', ``repetition'', and ``organization'', as discussed below:
\begin{j_itemize}
    \item[$\bullet$] \textbf{Meaning} refers to encoding information that holds significant importance, such as events that deeply impact individuals.
    \item[$\bullet$] \textbf{Association} involves pairing related objects with the encoded information, like remembering a ``bird'' based on characteristics such as flying.
    \item[$\bullet$] \textbf{Repetition} indicates the likelihood of information being transferred to the Long-term Store, as repeated exposure to scenes enhances memory retention.
    \item[$\bullet$] \textbf{Organization} suggests that the storage location in the Long-term Store is influenced by the content itself, such as recalling ``hamburger'' in the brain's ``food section''.
\end{j_itemize}

In summary, it is evident that memory storage, especially for long-term memory, is a relatively complex process. It involves processing perception, which we call ``encoding'', to produce information stored in the brain. The quality of memory storage largely determines the subsequent quality of memory retrieval and the rate of memory forgetting.

\subsubsection{Memory Retrieval}
\label{sec:human_memory:processing:retrieval}

According to Tulving et al.~\cite{tulving1973encoding}, memory retrieval involves a conscious search and identification of information stored in memory. They emphasized that this process is not merely the activation of learned associations or the recall of stored traces. Instead, it involves complex interactions between stored information and specific features of the current retrieval context. Providing cues similar to the original stimuli that formed the memory can enhance retrieval effectiveness. An example of this is individuals visiting a place they have previously stayed at, which is likely to evoke the events they experienced there. The hippocampus of brain is involved in the complex process of retrieving specific memory based on cues~\cite{marr1991simple,o2014complementary}.

During the 1970s, one of the most prominent theories of memory retrieval, coinciding with Tulving et al.'s research, was the Generation-Recognition Theory~\cite{shiffrin1969memory,underwood1969memory,tulving1973encoding}. 
This theory outlines the stages of memory retrieval, dividing them into the generation stage and the recognition stage. 
In the generation stage, individuals attempt to implicitly generate possible responses from memory, such as words, images, or other information associated with given cues. 
The recognition stage involves evaluating the responses generated in the generation stage to determine whether they meet specific acceptance criteria. In the recognition stage, the generated responses are matched against stored memory traces to see if they align with the specific information that was originally learned. Recognition requires not only identifying whether the response is familiar, but also confirming its context and correctness. For instance, it involves determining whether a recalled fact was encountered in a specific learning context or whether an image accurately represents an event experienced previously. It is concluded that the effectiveness of retrieval is enhanced if the process provides cues similar to those used during encoding~\cite{tulving1973encoding,bower1972encoding}.

\subsubsection{Memory Forgetting}
\label{sec:human_memory:processing:forgetting}

Memory forgetting is commonly perceived as the erasure of memory from storage. However, Atkinson et al.~\cite{shiffrin1969memory} proposed in 1969 that memory forgetting and related phenomena are often not due to the disappearance of information from long-term storage, but rather failures in the retrieval process. As individuals are exposed to increasing amounts of external information, their long-term memory grows correspondingly vast. The expanding scope of memory retrieval can cause interference between similar memory, leading to retrieval failures. 
This interference leads to the phenomenon known as ``memory forgetting'', which is a form of passive forgetting. 
On the other hand, some studies~\cite{hardt2013decay,medina2018neural,costanzi2021forgetting,anderson2021active} suggested that in addition to passive forgetting, humans can also achieve active regulation of the memory system through a mechanism that suppresses redundant information to facilitate the storage and retrieval of key memory.
This type of forgetting mechanism is termed ``active forgetting''.

The systematic study of memory forgetting dates back to the late 19th century. In 1885, Ebbinghaus et al.~\cite{ebbinghaus1885memory} demonstrated that 38 repetitions over three days had a similar effect to 68 repetitions in one day. In 1897, Jost's Law ~\cite{jost1897law} indicated that the probability of forgetting distant memory is higher than that of recent memory. Later, in 1957, Underwood et al.~\cite{underwood1957interference} explained a phenomenon where previously learned or concurrently learned information affects the forgetting of specific memory, which is termed ``interference''. These studies elucidate factors that can be utilized to reduce the probability of memory forgetting, including long intervals, high recency, and low interference, providing insights into effective strategies for retaining key information. For instance, leveraging the three strategies, individuals can improve word test results by reviewing words over extended intervals instead of repeating them in a short period (long intervals), focusing on word review as tests approach (high recency), and using specific markers to distinguish easily confused words (low interference). Moreover, there are active mechanisms in the brain to alleviate memory forgetting. Some studies suggest that the hippocampus engages in memory replay during certain phases, such as sleep, to consolidate memory and prevent forgetting. For example, place cells in the hippocampus spontaneously replay past trajectories to strengthen spatial memory~\cite{olafsdottir2018role}. Additionally, coordinated activity between the hippocampus and the visual cortex supports the replay of episodic memory, further aiding memory consolidation~\cite{ji2007coordinated}.

Studies that employ specific functions to describe memory forgetting curves are also crucial to our understanding of memory patterns. In 1970, Wickelgren et al.~\cite{wickelgren1970time} demonstrated through a letter memory experiment that the strength of memory traces in short-term memory (working memory) decays with increasing delay and/or interference, following an exponential form. Similarly, in 1985, White et al.~\cite{white1985characteristics} found that the memory forgetting curve in pigeons' discriminative ability to sample stimuli under different delay conditions could be well described by a simple negative exponential function. 
This function has two parameters: one for the initial discriminative ability and another for the rate at which this ability decreases with increasing delay intervals. However, in 1997, Wixted et al.~\cite{wixted1997genuine} conducted a series of experiments showing that forgetting curves are better described by a power function than by an exponential function, regardless of whether arithmetic or geometric averaging is used. Nevertheless, Anderson et al.~\cite{anderson1997memory} argued that the arithmetic averaging of exponential memory curves might produce an artifactual power memory curve. These findings highlight the complexity of accurately modeling memory forgetting curves and suggest that further research is needed to resolve these discrepancies.

In summary, memory forgetting can be caused by interference between memory that leads to retrieval failure; active forgetting of redundant information can facilitate the storage and retrieval of critical memory; long intervals, high recency, and low interference reduce the probability of memory forgetting; the memory replay mechanism of the hippocampus contributes to the alleviation of memory forgetting; and the quantitative modeling of memory forgetting helps researchers understand the patterns of memory.

\subsection{Summary}

Human memory exhibits a complex hierarchy, along with storage, retrieval, and forgetting mechanisms.
External stimuli briefly stay in the sensory register, with some information transitioning to short-term memory (working memory). Subsequently, information can be stored in long-term memory, forming more enduring memory. Information in long-term memory can be retrieved based on specific cues, playing a crucial role in using personal experiences, general knowledge to solve daily problems, and mastering skills such as riding a bicycle.
Human memory, particularly long-term memory, also has its flaws, such as forgetting due to memory interference.
Understanding human memory, especially long-term memory, from the perspectives of brain science and cognitive science can help us to better understand AI systems that employ similar mechanisms.

Additionally, some research indicates that human long-term memory cannot adapt its processing mechanisms through environmental changes and that this adjustment is only achievable through evolution~\cite{sweller2003evolution,manns2006evolution,allen2013evolution}. The evolutionary traits of human long-term memory that aid in survival are not necessarily well-suited for handling complex tasks in modern society, such as quickly grasping intricate knowledge for exams.
Addressing similar limitations in AI long-term memory systems presents significant research value.

\begin{figure}
\centering
\includegraphics[scale=0.45]{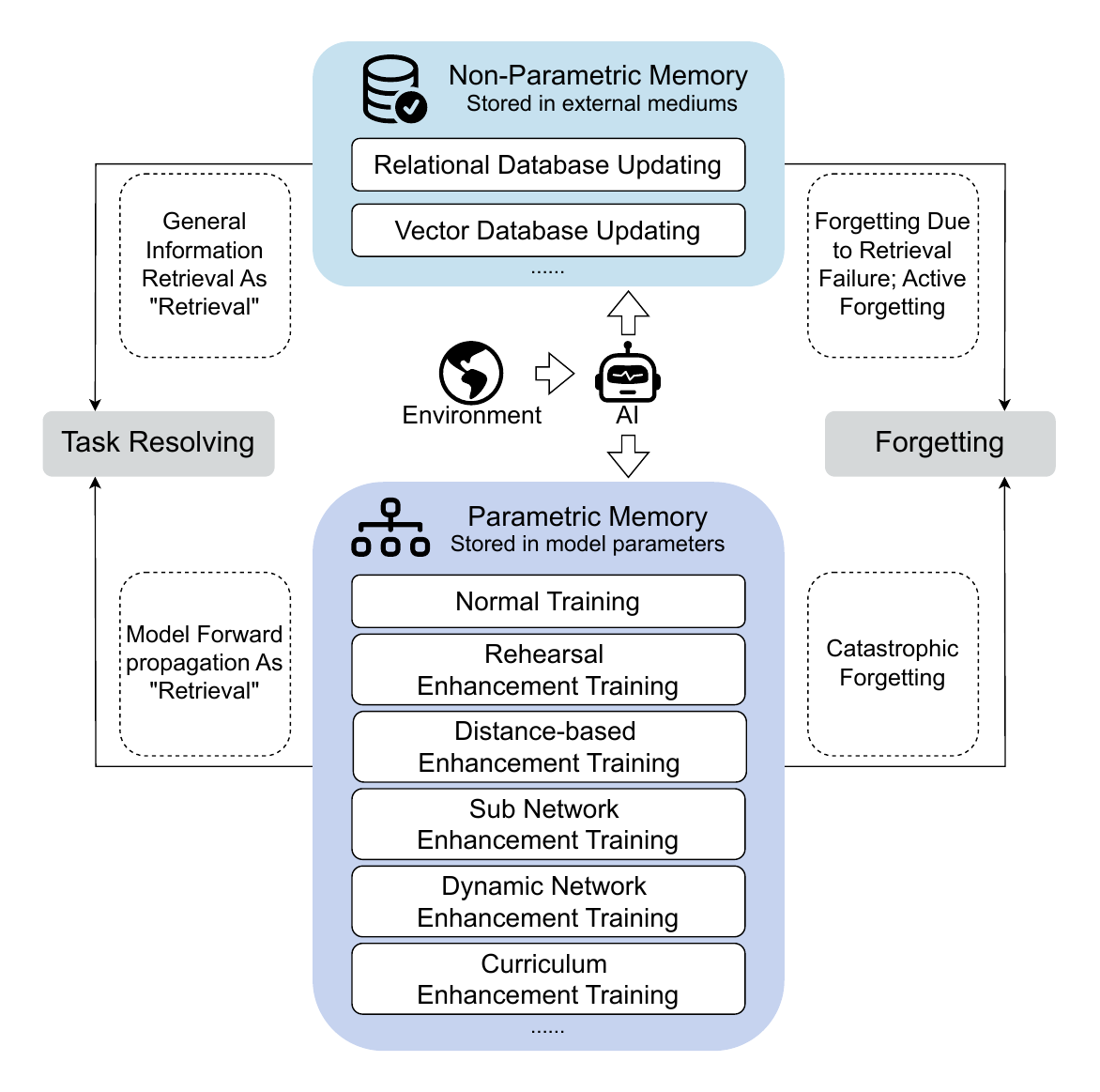}
\caption{Diagram of Storage, Retrieval, and Forgetting of non-parametric memory and parametric memory. In the diagram, \textbf{Relational Database Updating} and \textbf{Vector Database Updating} within the Non-Parametric Memory module can be referenced in Sec.~\ref{sec:AI_memory:non_parametric:memory_storage}; \textbf{Rehearsal Enhancement Training}, \textbf{Distance-based Enhancement Training}, \textbf{Sub Network Enhancement Training}, \textbf{Dynamic Network Enhancement Training} and \textbf{Curriculum Enhancement Training} within the Parametric Memory module can be referenced in Sec.~\ref{sec:AI_memory:parametric:memory_forgetting}.}
\label{fig:foundation}
\end{figure}

\definecolor{NodeColor}{HTML}{D5D8D9}
\definecolor{NodeColor1}{HTML}{C6D3EF}
\definecolor{NodeColor2}{HTML}{C6E1EF}

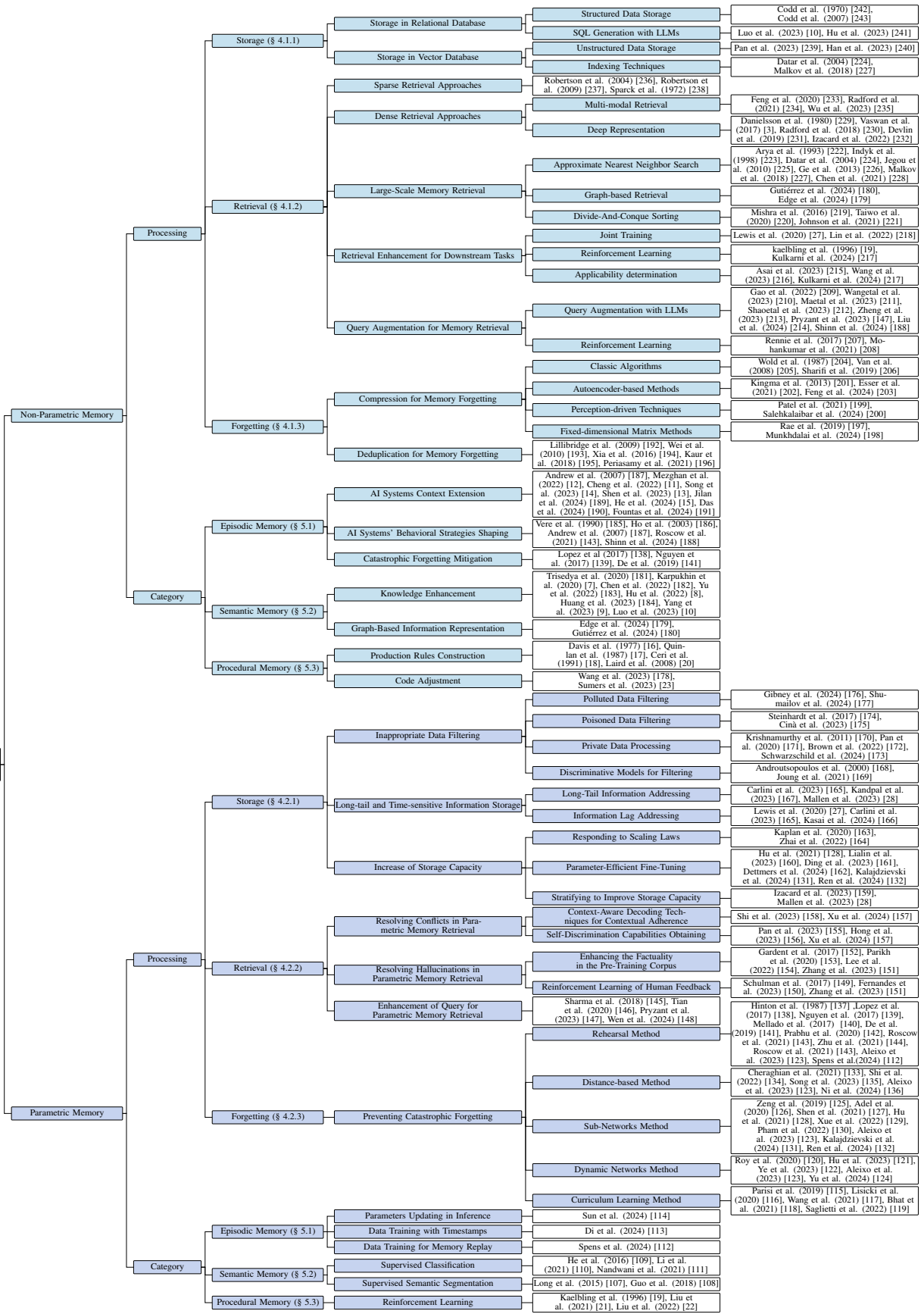
\begin{figure*}[!hp]
    \centering

\tikzset{
    basic/.style = {draw, text width=3cm, align=center, font=\sffamily, rectangle},
    root/.style = {basic, minimum height=3cm, rounded corners=2pt, thick, align=center, text width=10cm, font=\fontsize{35pt}{35pt}\selectfont\bfseries},
    onode/.style = {basic, minimum height=1.2cm, rounded corners=2pt, align=center, fill=NodeColor,text width=10cm, font=\fontsize{23.5pt}{23.5pt}\selectfont},
    parametricnode/.style = {basic, minimum height=1.2cm, rounded corners=2pt, align=center, fill=NodeColor1,text width=10cm, font=\fontsize{23.5pt}{23.5pt}\selectfont},
    nonparametricnode/.style = {basic, minimum height=1.2cm, rounded corners=2pt, align=center, fill=NodeColor2,text width=10cm, font=\fontsize{23.5pt}{23.5pt}\selectfont},
    wnode/.style = {basic, minimum height=1.2cm, rounded corners=2pt, align=center, fill=NodeColor,text width=6cm, font=\fontsize{23.5pt}{23.5pt}\selectfont},
    parametricwnode/.style = {basic, minimum height=1.2cm, rounded corners=2pt, align=center, fill=NodeColor1,text width=6cm, font=\fontsize{23.5pt}{23.5pt}\selectfont},
    nonparametricwnode/.style = {basic, minimum height=1.2cm, rounded corners=2pt, align=center, fill=NodeColor2,text width=6cm, font=\fontsize{23.5pt}{23.5pt}\selectfont},
    ynode/.style = {basic, minimum height=1.2cm, rounded corners=2pt, align=center, fill=NodeColor,text width=11cm, font=\fontsize{23.5pt}{23.5pt}\selectfont},
    parametricynode/.style = {basic, minimum height=1.2cm, rounded corners=2pt, align=center, fill=NodeColor1,text width=10cm, font=\fontsize{23.5pt}{23.5pt}\selectfont},
    nonparametricynode/.style = {basic, minimum height=1.2cm, rounded corners=2pt, align=center, fill=NodeColor2,text width=10cm, font=\fontsize{23.5pt}{23.5pt}\selectfont},
    xnode/.style = {basic, minimum height=1.2cm, rounded corners=2pt, align=center, fill=NodeColor,text width=17cm, font=\fontsize{23.5pt}{23.5pt}\selectfont},
    parametricxnode/.style = {basic, minimum height=1.2cm, rounded corners=2pt, align=center, fill=NodeColor1,text width=17cm, font=\fontsize{23.5pt}{23.5pt}\selectfont},
    nonparametricxnode/.style = {basic, minimum height=1.2cm, rounded corners=2pt, align=center, fill=NodeColor2,text width=17cm, font=\fontsize{23.5pt}{23.5pt}\selectfont},
    tnode/.style = {basic, minimum height=1.2cm, rounded corners=2pt, align=left, text width=17cm, align=center, font=\fontsize{23.5pt}{23.5pt}\selectfont},
    edge from parent/.style={draw=black, solid}
}
\resizebox{1\textwidth}{!}{
    \begin{forest} 
      for tree={
        grow=east,
        parent anchor=east,
        child anchor=west,
        anchor=center,
        edge path={
          \noexpand\path [draw, \forestoption{edge}] (!u.parent anchor) -- +(10pt,0) |- (.child anchor)\forestoption{edge label};
        },
    }
      [AI Long-term Memory, root, l sep=10mm, 
        [Parametric Memory, parametricnode,  l sep=10mm,
          [Category, parametricwnode,  l sep=10mm,
            [Procedural Memory (§~\ref{sec:hierarchy_ai_memory:procedural}), parametricynode,  l sep=10mm,
              [Reinforcement Learning, parametricxnode,  l sep=10mm,
                [{Kaelbling et al. (1996)~\cite{kaelbling1996reinforcement}, Liu et al. (2021)~\cite{liu2021automated}, Liu et al. (2022)~\cite{liu2022feature}}, tnode]
              ]
            ]
            [Semantic Memory (§~\ref{sec:hierarchy_ai_memory:semantic}), parametricynode,  l sep=10mm,
              [Supervised Semantic Segmentation, parametricxnode,  l sep=10mm,
                [{Long et al. (2015)~\cite{long2015fully}, Guo et al. (2018)~\cite{guo2018review}}, tnode]
              ]
              [Supervised Classification, parametricxnode,  l sep=10mm,
                [{He et al. (2016)~\cite{he2016deep}, Li et al. (2021)~\cite{li2021survey}, Nandwani et al. (2021)~\cite{nandwani2021review}}, tnode]
              ]
            ]
            [Episodic Memory (§~\ref{sec:hierarchy_ai_memory:episodic}), parametricynode,  l sep=10mm,
              [Data Training for Memory Replay, parametricxnode, l sep=10mm, 
                [Spens et al. (2024)~\cite{spens2024generative}, tnode]
              ]
              [Data Training with Timestamps, parametricxnode,  l sep=10mm,
                [Di et al. (2024)~\cite{di2024grounded}, tnode]
              ]
              [Parameters Updating in Inference, parametricxnode,  l sep=10mm,
                [Sun et al. (2024)~\cite{sun2024learning}, tnode]
              ]
            ]
          ]
          [Processing, parametricwnode,  l sep=10mm,
            [Forgetting (§~\ref{sec:AI_memory:parametric:memory_forgetting}), parametricynode,  l sep=10mm,
              [Preventing Catastrophic Forgetting, parametricxnode,  l sep=10mm,
                [Curriculum Learning Method, parametricxnode,  l sep=10mm,
                  [{Parisi et al. (2019)~\cite{parisi2019continual}, Lisicki et al. (2020)~\cite{lisicki2020evaluating}, Wang et al. (2021)~\cite{wang2021survey}, Bhat et al. (2021)~\cite{bhat2021cilea}, Saglietti et al. (2022)~\cite{saglietti2022analytical}}, tnode]
                ]
                [Dynamic Networks Method, parametricxnode,  l sep=10mm,
                  [{Roy et al. (2020)~\cite{roy2020tree}, Hu et al. (2023)~\cite{hu2023dense}, Ye et al. (2023)~\cite{ye2023self}, Aleixo et al. (2023)~\cite{aleixo2023catastrophic}, Yu et al. (2024)~\cite{yu2024boosting}}, tnode]
                ]
                [Sub-Networks Method, parametricxnode,  l sep=10mm,
                  [{Zeng et al. (2019)~\cite{zeng2019continual}, Adel et al. (2020)~\cite{Adel2020Continual}, Shen et al. (2021)~\cite{shen2021generative}, Hu et al. (2021)~\cite{hu2021lora}, Xue et al. (2022)~\cite{xue2022meta}, Pham et al. (2022)~\cite{pham2022continual}, Aleixo et al. (2023)~\cite{aleixo2023catastrophic}, Kalajdzievski et al. (2024)~\cite{kalajdzievski2024scaling}, Ren et al. (2024)~\cite{ren2024analyzing}}, tnode]
                ]
                [Distance-based Method, parametricxnode,  l sep=10mm,
                  [{Cheraghian et al. (2021)~\cite{cheraghian2021semantic}, Shi et al. (2022)~\cite{shi2022mimicking}, Song et al. (2023)~\cite{song2023learning}, Aleixo et al. (2023)~\cite{aleixo2023catastrophic}, Ni et al. (2024)~\cite{ni2024enhancing}}, tnode]
                ]
                [Rehearsal Method, parametricxnode,  l sep=10mm,
                  [{Hinton et al. (1987)~\cite{hinton1987using} ,Lopez et al. (2017)~\cite{lopez2017gradient}, Nguyen et al. (2017)~\cite{nguyen2018variational}, Mellado et al. (2017) ~\cite{mellado2017pseudorehearsal}, De et al. (2019)~\cite{de2019episodic}, Prabhu et al. (2020)~\cite{prabhu2020gdumb}, Roscow et al. (2021)~\cite{roscow2021learning}, Zhu et al. (2021)~\cite{zhu2021prototype}, Roscow et al. (2021)~\cite{roscow2021learning}, Aleixo et al. (2023)~\cite{aleixo2023catastrophic}, Spens et al.(2024)~\cite{spens2024generative}}, tnode]
                ]
              ]
            ]
            [Retrieval (§~\ref{sec:AI_memory:parametric:memory_retrieval}), parametricynode,  l sep=10mm,
              [Enhancement of Query for Parametric Memory Retrieval, parametricxnode,  l sep=10mm,
                [{Sharma et al. (2018)~\cite{sharma2018classification}, Tian et al. (2020)~\cite{tian2020deep}, Pryzant et al. (2023)~\cite{pryzant2023automatic}, Wen et al. (2024)~\cite{wen2024hard}}, tnode]
              ]
              [Resolving Hallucinations in Parametric Memory Retrieval, parametricxnode,  l sep=10mm,
                [Reinforcement Learning of Human Feedback, parametricxnode,  l sep=10mm,
                  [{Schulman et al. (2017)~\cite{schulman2017proximal}, Fernandes et al. (2023)~\cite{fernandes2023bridging}, Zhang et al. (2023)~\cite{zhang2023siren}}, tnode]
                ]
                [Enhancing the Factuality in the Pre-Training Corpus, parametricxnode,  l sep=10mm,
                  [{Gardent et al. (2017)~\cite{gardent2017creating}, Parikh et al. (2020)~\cite{parikhetal2020totto}, Lee et al. (2022)~\cite{lee2022factuality}, Zhang et al. (2023)~\cite{zhang2023siren}}, tnode]
                ]
              ]
              [Resolving Conflicts in Parametric Memory Retrieval, parametricxnode,  l sep=10mm,
                [Self-Discrimination Capabilities Obtaining, parametricxnode,  l sep=10mm,
                  [{Pan et al. (2023)~\cite{pan2023attacking}, Hong et al. (2023)~\cite{hong2023so}, Xu et al. (2024)~\cite{xu2024knowledge}}, tnode]
                ]
                [Context-Aware Decoding Techniques for Contextual Adherence, parametricxnode,  l sep=10mm,
                  [{Shi et al. (2023)~\cite{shi2023trusting}, Xu et al. (2024)~\cite{xu2024knowledge}}, tnode]
                ]
              ]
            ]
            [Storage (§~\ref{sec:AI_memory:parametric:storage}), parametricynode,  l sep=10mm,
              [Increase of Storage Capacity, parametricxnode,  l sep=10mm,
                [Stratifying to Improve Storage Capacity, parametricxnode,  l sep=10mm,
                  [{Izacard et al. (2023)~\cite{izacard2023atlas}, Mallen et al. (2023)~\cite{mallen2023trust}}, tnode]
                ]
                [Parameter-Efficient Fine-Tuning, parametricxnode,  l sep=10mm,
                  [{Hu et al. (2021)~\cite{hu2021lora}, Lialin et al. (2023)~\cite{lialin2023scaling}, Ding et al. (2023)~\cite{ding2023parameter}, Dettmers et al. (2024)~\cite{dettmers2024qlora}, Kalajdzievski et al. (2024)~\cite{kalajdzievski2024scaling}, Ren et al. (2024)~\cite{ren2024analyzing}}, tnode]
                ]
                [Responding to Scaling Laws, parametricxnode,  l sep=10mm,
                  [{Kaplan et al. (2020)~\cite{kaplan2020scaling}, Zhai et al. (2022)~\cite{zhai2022scaling}}, tnode]
                ]
              ]
              [Long-tail and Time-sensitive Information Storage, parametricxnode,  l sep=10mm,
                [Information Lag Addressing, parametricxnode,  l sep=10mm,
                  [{Lewis et al. (2020)~\cite{lewis2020retrieval}, Carlini et al. (2023)~\cite{carlini2023quantifying}, Kasai et al. (2024)~\cite{kasai2024realtime}}, tnode]
                ]
                [Long-Tail Information Addressing, parametricxnode,  l sep=10mm,
                  [{Carlini et al. (2023)~\cite{carlini2023quantifying}, Kandpal et al. (2023)~\cite{kandpal2023large}, Mallen et al. (2023)~\cite{mallen2023trust}}, tnode]
                ]
              ]
              [Inappropriate Data Filtering, parametricxnode,  l sep=10mm,
                [Discriminative Models for Filtering, parametricxnode,  l sep=10mm,
                  [{Androutsopoulos et al. (2000)~\cite{androutsopoulos2000experimental}, Joung et al. (2021)~\cite{joung2021automated}}, tnode]
                ]
                [Private Data Processing, parametricxnode,  l sep=10mm,
                  [{Krishnamurthy et al. (2011)~\cite{krishnamurthy2011privacy}, Pan et al. (2020)~\cite{pan2020privacy}, Brown et al. (2022)~\cite{brown2022does}, Schwarzschild et al. (2024)~\cite{schwarzschild2024rethinking}}, tnode]
                ]
                [Poisoned Data Filtering, parametricxnode,  l sep=10mm,
                  [{Steinhardt et al. (2017)~\cite{steinhardt2017certified}, Cin{\`a} et al. (2023)~\cite{cina2023wild}}, tnode]
                ]
                [Polluted Data Filtering, parametricxnode,  l sep=10mm,
                  [{Gibney et al. (2024)~\cite{gibney2024ai}, Shumailov et al. (2024)~\cite{shumailov2024ai}}, tnode]
                ]
              ]
            ]
          ]
        ]
        [Non-Parametric Memory, nonparametricnode,  l sep=10mm,
          [Category, nonparametricwnode,  l sep=10mm,
            [Procedural Memory (§~\ref{sec:hierarchy_ai_memory:procedural}), nonparametricynode,  l sep=10mm,
              [Code Adjustment, nonparametricxnode,  l sep=10mm,
                [{Wang et al. (2023)~\cite{wang2023voyager}, Sumers et al. (2023)~\cite{sumers2023cognitive}}, tnode]
              ]
              [Production Rules Construction, nonparametricxnode,  l sep=10mm,
                [{Davis et al. (1977)~\cite{davis1977production}, Quinlan et al. (1987)~\cite{quinlan1987generating}, Ceri et al. (1991)~\cite{ceri1991deriving}, Laird et al. (2008)~\cite{laird2008extending}}, tnode]
              ]
            ]
            [Semantic Memory (§~\ref{sec:hierarchy_ai_memory:semantic}), nonparametricynode,  l sep=10mm,
              [Graph-Based Information Representation, nonparametricxnode,  l sep=10mm,
                [{Edge et al. (2024)~\cite{edge2024local}, Guti{\'e}rrez et al. (2024)~\cite{gutierrez2024hipporag}}, tnode]
              ]
              [Knowledge Enhancement, nonparametricxnode,  l sep=10mm,
                [{Trisedya et al. (2020)~\cite{trisedya2020sentence}, Karpukhin et al. (2020)~\cite{qa2020karpukhin}, Chen et al. (2022)~\cite{wenhu2022reimage}, Yu et al. (2022)~\cite{yu2022survey}, Hu et al. (2022)~\cite{xixin2022gmtkbqa}, Huang et al. (2023)~\cite{rongjie2023makeanaudio}, Yang et al. (2023)~\cite{yang2023care}, Luo et al. (2023)~\cite{luo2023chatkbqa}}, tnode]
              ]
            ]
            [Episodic Memory (§~\ref{sec:hierarchy_ai_memory:episodic}), nonparametricynode,  l sep=10mm,
              [Catastrophic Forgetting Mitigation, nonparametricxnode, l sep=10mm,
                [{Lopez et al (2017)~\cite{lopez2017gradient}, Nguyen et al. (2017)~\cite{nguyen2018variational}, De et al. (2019)~\cite{de2019episodic}}, tnode]
              ]
              [AI Systems' Behavioral Strategies Shaping, nonparametricxnode,  l sep=10mm,
                [{Vere et al. (1990)~\cite{vere1990basic}, Ho et al. (2003)~\cite{ho2003comparing}, Andrew et al. (2007)~\cite{andrew2007episodic}, Roscow et al. (2021)~\cite{roscow2021learning}, Shinn et al. (2024)~\cite{shinn2024reflexion}}, tnode]
              ]
              [AI Systems Context Extension, nonparametricxnode,  l sep=10mm,
                [{Andrew et al. (2007)~\cite{andrew2007episodic}, Mezghan et al. (2022)~\cite{mezghan2022memory}, Cheng et al. (2022)~\cite{cheng2022xmem}, Song et al. (2023)~\cite{song2023moviechat}, Shen et al. (2023)~\cite{shen2023encode}, Jilan et al. (2024)~\cite{jilan2024egoInstructor}, He et al. (2024)~\cite{he2024ma}, Das et al. (2024)~\cite{das2024larimar}, Fountas et al. (2024)~\cite{fountas2024human}}, tnode]
              ]
            ]
          ]
          [Processing, nonparametricwnode,  l sep=10mm,
            [Forgetting (§~\ref{sec:AI_memory:non_parametric:memory_forgetting}), nonparametricynode,  l sep=10mm,
              [Deduplication for Memory Forgetting, nonparametricxnode,  l sep=10mm,
                  [{Lillibridge et al. (2009)~\cite{lillibridge2009sparse}, Wei et al. (2010)~\cite{wei2010mad2}, Xia et al. (2016)~\cite{xia2016comprehensive}, Kaur et al. (2018)~\cite{kaur2018data}, Periasamy et al. (2021)~\cite{periasamy2021efficient}}, tnode]
              ]
              [Compression for Memory Forgetting, nonparametricxnode,  l sep=10mm,
                [Fixed-dimensional Matrix Methods, nonparametricxnode,  l sep=10mm,
                  [{Rae et al. (2019)~\cite{rae2019compressive}, Munkhdalai et al. (2024)~\cite{munkhdalai2024leave}}, tnode]
                ]
                [Perception-driven Techniques, nonparametricxnode,  l sep=10mm,
                  [{Patel et al. (2021)~\cite{patel2021saliency}, Salehkalaibar et al. (2024)~\cite{salehkalaibar2024choice}}, tnode]
                ]
                [Autoencoder-based Methods, nonparametricxnode,  l sep=10mm,
                  [{Kingma et al. (2013)~\cite{kingma2013auto}, Esser et al. (2021)~\cite{esser2021taming}, Feng et al. (2024)~\cite{feng2024memristor}}, tnode]
                ]
                [Classic Algorithms, nonparametricxnode,  l sep=10mm,
                  [{Wold et al. (1987)~\cite{wold1987principal}, Van et al. (2008)~\cite{van2008visualizing}, Sharifi et al. (2019)~\cite{sharifi2019removing}}, tnode]
                ]
              ]
            ]
            [Retrieval (§~\ref{sec:AI_memory:non_parametric:memory_retrieval}), nonparametricynode,  l sep=10mm,
              [Query Augmentation for Memory Retrieval, nonparametricxnode,  l sep=10mm,
                [Reinforcement Learning, nonparametricxnode,  l sep=10mm,
                  [{Rennie et al. (2017)~\cite{rennie2017self}, Mohankumar et al. (2021)~\cite{mohankumar2021diversity}}, tnode]
                ]
                [Query Augmentation with LLMs, nonparametricxnode,  l sep=10mm,
                  [{Gao et al. (2022)~\cite{gao2023precise}, Wangetal et al. (2023)~\cite{wangetal2023query2doc}, Maetal et al. (2023)~\cite{maetal2023query}, Shaoetal et al. (2023)~\cite{shaoetal2023enhancing}, Zheng et al. (2023)~\cite{zheng2023take}, Pryzant et al. (2023)~\cite{pryzant2023automatic}, Liu et al. (2024)~\cite{liu2024query}, Shinn et al. (2024)~\cite{shinn2024reflexion}}, tnode]
                ]
              ]
              [Retrieval Enhancement for Downstream Tasks, nonparametricxnode,  l sep=10mm,
                [Applicability determination, nonparametricxnode,  l sep=10mm,
                  [{Asai et al. (2023)~\cite{asai2023self}, Wang et al. (2023)~\cite{wang2023skr}, Kulkarni et al. (2024)~\cite{kulkarni2024reinforcement}}, tnode]
                ]
                [Reinforcement Learning, nonparametricxnode,  l sep=10mm,
                  [{kaelbling et al. (1996)~\cite{kaelbling1996reinforcement}, Kulkarni et al. (2024)~\cite{kulkarni2024reinforcement}}, tnode]
                ]
                [Joint Training, nonparametricxnode,  l sep=10mm,
                  [{Lewis et al. (2020)~\cite{lewis2020retrieval}, Lin et al. (2022)~\cite{lin2022okvqa}}, tnode]
                ]
              ]
              [Large-Scale Memory Retrieval, nonparametricxnode,  l sep=10mm,
                [Divide-And-Conque Sorting, nonparametricxnode,  l sep=10mm,
                  [{Mishra et al. (2016)~\cite{mishra2016divide}, Taiwo et al. (2020)~\cite{taiwo2020comparative}, Johnson et al. (2021)~\cite{billion2021johnson}}, tnode]
                ]
                [Graph-based Retrieval, nonparametricxnode,  l sep=10mm,
                  [{Guti{\'e}rrez et al. (2024)~\cite{gutierrez2024hipporag}, Edge et al. (2024)~\cite{edge2024local}}, tnode]
                ]
                [Approximate Nearest Neighbor Search, nonparametricxnode,  l sep=10mm,
                  [{Arya et al. (1993)~\cite{arya1993approximate}, Indyk et al. (1998)~\cite{indyk1998approximate}, Datar et al. (2004)~\cite{datar2004locality}, Jegou et al. (2010)~\cite{jegou2010product}, Ge et al. (2013)~\cite{ge2013optimized}, Malkov et al. (2018)~\cite{malkov2018efficient}, Chen et al. (2021)~\cite{spann2021chen}}, tnode]
                ]
              ]
              [Dense Retrieval Approaches, nonparametricxnode,  l sep=10mm,
                [Deep Representation, nonparametricxnode,  l sep=10mm,
                  [{Danielsson et al. (1980)~\cite{danielsson1980euclidean}, Vaswan et al. (2017)~\cite{vaswani2017attention}, Radford et al. (2018)~\cite{radford2018improving}, Devlin et al. (2019)~\cite{kenton2019bert}, Izacard et al. (2022)~\cite{izacard2022unsupervised}}, tnode]
                ]
                [Multi-modal Retrieval, nonparametricxnode,  l sep=10mm,
                  [{Feng et al. (2020)~\cite{codebert2020zhangyin}, Radford et al. (2021)~\cite{clip2021radford}, Wu et al. (2023)~\cite{yusong2023audio}}, tnode]
                ]
              ]
              [Sparse Retrieval Approaches, nonparametricxnode,  l sep=10mm,
                  [{Robertson et al. (2004)~\cite{robertson2004simple}, Robertson et al. (2009)~\cite{bm252009robertson}, Sparck et al. (1972)~\cite{sparck1972statistical}}, tnode]
              ]
            ]
            [Storage (§~\ref{sec:AI_memory:non_parametric:memory_storage}), nonparametricynode,  l sep=10mm,
              [Storage in Vector Database, nonparametricxnode,  l sep=10mm,
                [Indexing Techniques, nonparametricxnode,  l sep=10mm,
                  [{Datar et al. (2004)~\cite{datar2004locality}, Malkov et al. (2018)~\cite{malkov2018efficient}}, tnode]
                ]
                [Unstructured Data Storage, nonparametricxnode,  l sep=10mm,
                  [{Pan et al. (2023)~\cite{pan2023survey}, Han et al. (2023)~\cite{han2023comprehensive}}, tnode]
                ]
              ]
              [Storage in Relational Database, nonparametricxnode,  l sep=10mm,
                [SQL Generation with LLMs, nonparametricxnode,  l sep=10mm,
                  [{Luo et al. (2023)~\cite{luo2023chatkbqa}, Hu et al. (2023)~\cite{hu2023chatdb}}, tnode]
                ]
                [Structured Data Storage, nonparametricxnode,  l sep=10mm,
                  [{Codd et al. (1970)~\cite{codd1970relational}, Codd et al. (2007)~\cite{codd2007relational}}, tnode]
                ]
              ]
            ]
          ]
        ]
      ]
    \end{forest}
 }
\caption{Taxonomy of AI Long-term Memory and a collection of representative related works. AI long-term memory can be divided into non-parametric memory and parametric memory based on whether it is stored within model parameters. Both of these methods have specific mechanisms for storage, retrieval, and forgetting (Sec.~\ref{sec:founddation_ai_memory}), and they are strongly related to human episodic, semantic, and procedural memory (Sec.~\ref{sec:hierarchy_ai_memory}).}
\label{fig:lit_surv}
\end{figure*}

\section{Long-term Memory of AI: on Storage Formats}
\label{sec:founddation_ai_memory}

In the previous sections, we examined the hierarchy and processing methods related to long-term memory in the human brain. Long-term memory, however, is not exclusive to humans; AI also possesses long-term memory mechanisms. AI long-term memory serves various purposes: it records episodic information from past events~\cite{jilan2024egoInstructor,andrew2007episodic}, learns semantic information~\cite{wenhu2022reimage,yang2023care}, and gains experience from observations or feedback~\cite{shinn2024reflexion,zhao2024expel}. AI long-term memory mirrors certain aspects of human long-term memory, including storage, retrieval, and forgetting processes; some AI long-term memory research is directly inspired by human long-term memory~\cite{wang2016semantic,cheng2022xmem,song2023moviechat,das2024larimar,fountas2024human}, while other studies align with human long-term memory processing mechanisms without intentional imitation.

Akin to the human brain, AI can process input stimuli to form long-term memory and schedule them appropriately. For instance, training of neural network models~\cite{lu2007survey,he2016deep} serves as a process of forming long-term memory of the models. During training, the model adjusts its weights through gradient descent~\cite{ruder2016overview}, while during inference, the updated weights are utilized to compute the model's output, such as predicting the category of an image~\cite{he2016deep,li2021survey}. If the model and training method are properly applied, the updated weights compared to the original weights can enable the model to better predict image categories. This process is analogous to how humans learn from images to store long-term memory (Sec.~\ref{sec:human_memory:processing:storage}) and retrieve these long-term memory at appropriate times to recognize specific images (Sec.~\ref{sec:human_memory:processing:retrieval}).

The aforementioned example shows that AI long-term memory can be implicitly stored in the parameters of AI models~\cite{petroni2019language,robertsetal2020much}. 
On the other hand, ome AI systems store long-term memory in external storage medias, such as databases, outside the AI models~\cite{hu2023chatdb,zhang2023long,shen2023encode}. 
Such long-term memory can be stored in its raw form, like text notes or structured labels, in format of relational model~\cite{hu2023chatdb,luo2023chatkbqa}, enabling AI systems to retrieve and utilize this information when needed to support it in performing tasks;
memory stored in external media can also take the form of vectors~\cite{zhang2023long,he2024g}, which more closely resembles the specific encoding process of human brains~\cite{tulving1973encoding} (Sec.~\ref{sec:human_memory:processing:storage}).

Based on these observations, we categorize AI long-term memory into non-parametric memory (Sec.~\ref{sec:AI_memory:non_parametric}) and parametric memory (Sec.~\ref{sec:AI_memory:parametric}), following the criteria illustrated by Lewis et al.~\cite{lewis2020retrieval} and Mallen et al.~\cite{mallen2023trust} and depicted in Fig.~\ref{fig:lit_surv}. Specifically, non-parametric memory refers to AI long-term memory stored in external media, while parametric memory refers to AI long-term memory stored within the model's parameters. A key distinction is that the model parameters are updated during the storage process only in the case of parametric memory. The processing mechanisms of non-parametric and parametric memory are shown in Fig.~\ref{fig:foundation}. In the subsequent sections, we offer a comprehensive discussion on the categorization of AI's long-term memory, review relevant literature, and identify both key challenges and potential solutions. 
Recognizing the prominent role of (large) language models in related works, we highlight these models as illustrative examples of AI systems in certain sections, while maintaining a broader scope.

\subsection{Non-Parametric Memory}
\label{sec:AI_memory:non_parametric}

Non-parametric memory refers to long-term memory that is stored externally to the AI models. This type of memory can be retrieved based on specific cues when performing tasks. 
Retrieval-Augmented Generation (RAG)~\cite{gao2023retrieval,zhao2024retrieval} is an example framework that leverages non-parametric memory.
RAG retrieves information from external data sources (``non-parametric memory'') and produces responses that are informed by the retrieved information.
We explore the related work of storage, retrieval and forgetting of non-parametric memory in Sec.~\ref{sec:AI_memory:non_parametric:memory_storage}, Sec.~\ref{sec:AI_memory:non_parametric:memory_retrieval} and Sec.~\ref{sec:AI_memory:non_parametric:memory_forgetting}.

\subsubsection{Storage of Non-Parametric Memory} 
\label{sec:AI_memory:non_parametric:memory_storage}

Non-parametric memory can be stored in various media, such as databases, file systems, and computer memory. File systems are used by computer operating systems to efficiently store, organize, manage, and access data on disks~\cite{ghemawat2003google,bovet2005understanding}. Computer memory, on the other hand, provides temporary storage space for a computer~\cite{kaxiras2008computer,hennessy2011computer}. Among these storage options, databases are particularly noteworthy for their scalability and maintenance efficiency, making them the preferred medium for accessing non-parametric memory. 
The databases provide diverse storage options for different types of non-parametric memory. For example, sparse features derived from user interactions can be stored in relational databases~\cite{codd2007relational} for use in recommender systems~\cite{ko2022survey}, whereas vectorized document chunks can be stored in vector databases~\cite{pan2023survey,han2023comprehensive} for RAG systems~\cite{gao2023retrieval,zhao2024retrieval}. The following sections illustrate how these two types of database are utilized for storing non-parametric memory.

\begin{j_itemize}

    \item[$\bullet$] \textbf{Relational database.} Relational databases store data using a relational model~\cite{codd1970relational,codd2007relational}. For example, consider a structured data ``tiger'' - ``belongs to'' - ``feline''. In a relational database, this can be stored as a record in a table representing a collection of animals. These databases support encoding of structured non-parametric memory. Operations on data in relational databases can be performed using Structured Query Language (SQL)~\cite{beaulieu2009learning}. Several studies leveraged LLMs to generate SQL operations for storing, manipulating, and retrieving non-parametric memory in relational databases. For instance, Luo et al.~\cite{luo2023chatkbqa} utilized LLMs to generate \texttt{SELECT} SQL operations, which are used to select entries that satisfy given conditions, to retrieve long-term memory from a relational database. Similarly, Hu et al.~\cite{hu2023chatdb} used LLMs to convert the original queries to multi-step SQL commands to perform \texttt{INSERT} (insert a new entry), \texttt{UPDATE} (update an existing entry), \texttt{SELECT}, and \texttt{DELETE} (remove an existing entry) operations, which enables the flexible handling of non-parametric memory.
    
    \item[$\bullet$] \textbf{Vector Database.} Traditional relational databases require highly structured data and are not effective at data types like texts, images, and audios~\cite{jing2024large}. To address this limitation, vector databases have been developed to store vectors encoded from various types of data, making them more suitable for unified storage of these unstructured data types, and supports more efficient retrieval based on vector similarity~\cite{pan2023survey,han2023comprehensive}. Non-parametric memory suitable for storage in vector databases can be derived from representation vectors of different modalities (e.g., text, images) obtained through contrastive learning~\cite{clip2021radford,yusong2023audio,girdhar2023imagebind}, language representation vectors acquired via pre-training methods like masked word prediction and next sentence prediction~\cite{kenton2019bert,lan2019albert,liu2019roberta}, or key and value vectors from the attention mechanism in Transformer models~\cite{sukhbaatar2019augmenting,qiu2024empirical,wang2024augmenting,munkhdalai2024leave}. Moreover, compared to traditional indexing techniques in relational databases such as B-trees and hash tables, indexing techniques specialized for high-dimensional vectors can be adopted in vector databases, such as indexing of Locality-Sensitive Hashing (LSH)~\cite{datar2004locality} and Hierarchical Navigable Small World (HNSW) graphs~\cite{malkov2018efficient}. These indexing methods enable the efficient storage of non-parametric memory. Multiple studies have demonstrated the advantages of using vector databases to store non-parametric memory. For example, Zhang et al.~\cite{zhang2023long} utilized vector databases to provide personalized long-term memory for LLMs, which enhances the personalization of AI assistants. Shen et al.~\cite{shen2023encode} explored the use of vector databases to enhance the long video understanding capability of AI systems by encoding the natural language descriptions of video frames into vectors. 
    This can be considered as long-term memory within video contexts.
    
\end{j_itemize}

In summary, relational databases can be used to store structured data in non-parametric memory~\cite{codd1970relational,codd2007relational,luo2023chatkbqa,hu2023chatdb}, while vector databases are suitable for storing non-parametric memory of various modalities~\cite{pan2023survey,zhang2023long}. Considering the diversity of non-parametric memory, vector databases are often more appropriate for their storage.

\subsubsection{Retrieval of Non-Parametric Memory}
\label{sec:AI_memory:non_parametric:memory_retrieval}

Non-parametric memory functions as a collection of stored information, and its retrieval process is similar to the concept of information retrieval. The retrieval of non-parametric memory involves locating the most relevant $K$ memory fragments. This process consists of two main steps. It first calculates the relevance score between the query $q$ and each fragment $m$ in the non-parametric memory, and then ranks the fragments $m$ based on their relevance scores and selects the top-$K$ fragments $m$ with the highest scores to form the retrieval results~\cite{baeza1999modern,manning2008introduction,izacard2022unsupervised}.

Based on our review of existing work, we identify two primary retrieval methods suitable for non-parametric memory including sparse methods based on bag-of-words models and dense methods based on deep representation learning for vector encoding and relevance score calculation~\cite{arabzadeh2021predicting,luan2021sparse,izacard2022unsupervised,mandikal2024sparse}. Sparse retrieval, which requires both the query and the memory fragment to be in text format, involves methods such as BM25~\cite{robertson2004simple,bm252009robertson} and TF-IDF~\cite{sparck1972statistical}. In contrast, dense retrieval matches queries with memory fragments in deep semantic representations format~\cite{radford2018improving,kenton2019bert} that cater to various modalities, including text, images, audio and code~\cite{codebert2020zhangyin,clip2021radford,yusong2023audio}. In this case, memory fragments are typically stored in a vector database to enable fast retrieval. Dense retrieval methods often utilize deep models, such as Transformer-based models~\cite{vaswani2017attention}, to encode the query $q$ and the memory fragment $m$ into vectors of equal dimension. 
The ranking score can be obtained by calculating the Euclidean distance~\cite{danielsson1980euclidean} or by performing a dot product~\cite{izacard2022unsupervised} between the query vector and the memory fragment vector.

As dense retrieval supports deep semantic matching across various modalities, it emerges as a more effective method for non-parametric memory. 
In applying dense retrieval to non-parametric memory storage, we identify three key challenges along with potential solutions found in the literature: (1) efficiently managing large-scale memory retrieval, (2) improving retrieval effectiveness for downstream tasks, and (3) augmenting query for memory retrieval.

\begin{j_itemize}

    \item[$\bullet$] \textbf{Handling of Large-Scale Memory Retrieval.} Efficient retrieval from large-scale non-parametric memory, including the effective assessment of the matching scores between items and efficient sorting, poses a significant challenge. Approximate Nearest Neighbor Search (ANNS)~\cite{arya1993approximate,indyk1998approximate,malkov2018efficient,spann2021chen} is a commonly used method for efficiently retrieving large-scale non-parametric memory, effectively reducing computational complexity and finding results that are approximately accurate. ANNS can be efficiently realized through methods such as hashing functions that map similar vectors into the same bucket for focused searching~\cite{datar2004locality}, quantization techniques that decompose high-dimensional vectors into sub-vectors and utilize their centroids as matching objects for reducing computational complexity of matching score calculations~\cite{jegou2010product,ge2013optimized}, and hierarchical approaches that employ layered sparse graphs~\cite{malkov2018efficient} or clusters~\cite{spann2021chen} to implement localized search. Furthermore, using graph structures to construct non-parametric memory is highly effective for processing retrieval tasks from large-scale non-parametric memory, particularly for queries involving multiple entities~\cite{gutierrez2024hipporag,edge2024local}. For instance, consider the query ``In which of Antoine's books is the planet B612 mentioned?''. If there are only two relevant text fragments in text-based non-parametric memory, one stating ``The Little Prince mentions the planet B612'', and the other noting ``Antoine is the author of The Little Prince'', traditional retrieval might return all text fragments containing ``Antoine'' or ``planet B612'', causing too much redundant information in the results. In contrast, if this text-based non-parametric memory is represented in a graph format where entities and their relationships are clearly depicted, the entity of ``The Little Prince'' will receive the highest score due to its direct connections with the entities ``Antoine'' and ``planet B612'' mentioned in the query, thereby ensuring the accuracy of the retrieval results. Additionally, the divide-and-conquer strategy~\cite{mishra2016divide,taiwo2020comparative} offers an efficient approach for the sorting phase in retrieval of large-scale non-parametric memory. For example, Chen et al.~\cite{billion2021johnson} addressed the sorting of long list with relevant score by employing a recursive sorting function that can bisect the problem. They sort the entire list by recursively combining two separately sorted sub lists with a merge function.

    \item[$\bullet$] \textbf{Retrieval Enhancement for Downstream Tasks.} Non-Parametric Memory retrieval is often used to support downstream models in tasks such as question answering~\cite{qa2020karpukhin,xixin2022gmtkbqa,yang2023care,luo2023chatkbqa}. If the retrieval model and the downstream model are trained independently, it leads to suboptimal performance of the overall system. Considering this aspect, optimizing the retrieval model using feedback from the downstream model can be adopted to enhance the performance of both the retrieval process and the downstream task. Within the purview of our investigation, we delineate three principal solutions. The first approach is joint training, where the retrieval model and the downstream task model are trained together. For instance, Lewis et al.~\cite{lewis2020retrieval} jointly trained the retrieval model and the downstream text generating model by minimizing negative marginal log-likelihood for each target associated with the given input. Lin et al.~\cite{lin2022okvqa} proposed a loss function that integrates predictions from both models and a pseudo relevance score, which assesses whether the retrieved content contains the target answer, enabling joint training of the retrieval and question-answering models. 
    The second approach involves using reinforcement learning~\cite{kaelbling1996reinforcement,kulkarni2024reinforcement} to optimize the retrieval model. This method designs reward functions based on downstream performance, encouraging the retrieval model when the downstream task is well executed and discouraging it otherwise.
    The third approach allows the system to decide whether to invoke retrieval based on the context~\cite{asai2023self,wang2023skr,kulkarni2024reinforcement}. For example, the system developed by Wang et al.~\cite{wang2023skr} assesses enhancement brought to problem solving by external knowledge, and triggers the retrieval component only when the retrieval of external knowledge is deemed necessary.

    \item[$\bullet$] \textbf{Query Augmentation for Memory Retrieval.} 
    Query augmentation in information retrieval refers to techniques aimed at enhancing or modifying a user's query to improve the quality of search results. The main objective of query augmentation is to reformulate the user's initial query to better match the relevant documents in the database, effectively bridging the gap between the user's query and the information they seek. 
    The query augmentation method used in non-parametric memory retrieval is consistent with that in information retrieval.
    Recently, beyond traditional query augmentation methods~\cite{aljadda2014crowdsourced,abdelkawi2019complex}, LLM-based query augmentation techniques have gained attention. For example, Query2Doc~\cite{wangetal2023query2doc} generates pseudo-documents with relevant information through LLMs, expanding the original query by incorporating more details. 
    Step-Back Prompting~\cite{zheng2023take} employs LLMs to identify high-level terms or topics that capture the core concepts of a query, generating queries at various levels of abstraction to enhance search outcomes. Similarly, Hypothetical Document Embeddings (HyDE)~\cite{gao2023precise} convert the original query into a hypothetical document, retrieving information based on its encoded representation. Rewrite-Retrieve-Read~\cite{maetal2023query} trains the rewriter model for query rewriting based on samples of rewritten queries generated by LLMs that are accurately predicted. Iterative Retrieval-Generation Synergy (ITER-RETGEN)~\cite{shaoetal2023enhancing} iteratively rewrites the original query by using generated or retrieved answers, thus improving query effectiveness. 
    LLM-based techniques not limited to query augmentation also offer valuable concepts. For example, Reflectxion~\cite{shinn2024reflexion} enables LLMs to revise experiences based on environmental feedback, a mechanism that could be adapted for revising queries using similar feedback. Prompt Optimization with Textual Gradients (ProTeGi)~\cite{pryzant2023automatic} allows LLMs to optimize the original prompt by reflecting on the error between predicted and actual results, a process applicable to queries in retrieval contexts.
    Reinforcement learning offers another approach to query augmentation, optimizing models to enhance reward signals tied to task performance. For example, Self-Critical Sequence Training (SCST)~\cite{rennie2017self} normalizes rewards during test-time inference by focusing on sequences (which can include queries) that exceed current system performance, encouraging improvements. Close Variant Generation (CLOVER)~\cite{mohankumar2021diversity} utilizes diversity-driven reinforcement learning to optimize query rewriting, enhancing both the quality and diversity of query expansions.

\end{j_itemize}

In summary, the retrieval of non-parametric memory can be divided into sparse retrieval and dense retrieval. 
The former is based on the matching of textual fragments, while the latter is based on the matching of deep semantic representations. 
We identify three key challenges in the retrieval of non-parametric memory: achieving large-scale retrieval, enhancing retrieval for downstream tasks, and query augmentation. We discuss both existing and potential solutions to address these challenges.

\subsubsection{Forgetting of Non-Parametric Memory}
\label{sec:AI_memory:non_parametric:memory_forgetting}

The previous subsection mentioned that retrieving non-parametric memory involves two main steps: generating relevance scores and ranking to select the top-K fragments. When the retrieval model for generating relevance scores remains unchanged and the top-K value of selected fragments is constant, the probability of correctly selecting the target memory fragment decreases as the amount of information in non-parametric memory increases. This increase in information raises the likelihood of retrieval failure. 
According to the theory of human memory proposed by Atkinson et al.~\cite{shiffrin1969memory}, the phenomenon of retrieval failure can be considered a form of forgetting, which is also a key challenge in non-parametric memory systems. On the other hand, we can draw on the mechanism of human active forgetting (discussed in Sec.~\ref{sec:human_memory:processing:forgetting}) to actively lessen redundancy of non-parametric memory, thereby reducing storage space and improving retrieval efficiency. Implementing active forgetting of redundancy can alleviate the forgetting of critical information. We review the literature and
identify several practical solutions for implementing active forgetting:

\begin{j_itemize}

    \item[$\bullet$] \textbf{Compression for Memory Forgetting.} The goal of compression, especially the lossy compression, is to reduce the storage space or transmission bandwidth required to represent data, while ensuring that the data quality remains acceptable~\cite{lelewer1987data,salomon2002data,goyal2008compressive}. 
    Some classic feature dimension reduction algorithms can be used to achieve data compression, such as Principal Component Analysis (PCA)~\cite{wold1987principal}, which is used to reduce the dimension of features and identify the most important components within a dataset. For instance, Sharifi et al.~\cite{sharifi2019removing} used PCA to detect the less significant parts of features, effectively reducing storage redundancy by discarding unnecessary information. Another classic algorithm for data compression is t-Distributed Stochastic Neighbor Embedding (t-SNE)~\cite{van2008visualizing}, which achieves the mapping of high-dimensional data to low-dimensional space by optimizing an objective function that measures whether similar points in high-dimensional space remain similar in low-dimensional space. In contrast, autoencoders are a newer approach to data compression~\cite{kingma2013auto,feng2024memristor}. For example, Kingma et al.~\cite{kingma2013auto} proposed variational autoencoders that project high-dimensional data into a lower-dimensional latent space using an encoder, employ variational inference to approximate the posterior distribution of the data within the latent space. Improving autoencoders with vector quantization can more efficiently achieve data compression, such as the Vector-Quantized Generative Adversarial Network proposed by Esser et al.~\cite{esser2021taming}, which learns a discrete codebook to map continuous data latent representations to a finite number of code vectors. Data compression can also be achieved through a perception-driven approach. For example, Patel et al.~\cite{patel2021saliency} allocated more bits to key areas in images (such as faces and text) using a saliency mask, ensuring the visual quality of compressed images and achieving efficient image compression that aligns with human visual perception. Salehkalaibar et al.~\cite{salehkalaibar2024choice} explored perceptual compression techniques by comparing Perception Loss Function based on Joint Distribution (PLF-JD), which considers the joint distribution of video frames, and Perception Loss Function based on Framewise Marginal Distribution (PLF-FMD), focusing on the marginal distribution of individual frames. While PLF-JD excels at maintaining inter-frame coherence, it is more susceptible to error propagation. In contrast, PLF-FMD demonstrates superior capability in correcting errors across frames. Data compression can be applied in various applications, such as video processing, as demonstrated by Liu et al. in the literature~\cite{liu2020deep} and Bidwe et al. in the literature~\cite{bidwe2022deep}, where deep learning-based video compression techniques can reduce superfluous information and enhance the storage and retrieval of video data. Additionally, using a fixed-dimensional memory matrix to store an ever-expanding data source can be regarded as a form of data compression, where this compressed memory matrix can be retrieved through attention mechanism-based computations, reducing the complexity of storage and retrieval. In this regard, Rae et al.~\cite{rae2019compressive} achieve the extension of fixed-dimensional memory matrix by applying linear algebra components for compression, while Munkhdalai et al.~\cite{munkhdalai2024leave} update the memory matrix by overlaying the results of nonlinear key-value pair computations with retrieved memory content. These compression techniques can be used in active forgetting, where non-essential details are discarded to optimize storage and retrieval.

    \item[$\bullet$] \textbf{Deduplication for Memory Forgetting.} Data deduplication aims to reduce redundant information by identifying and eliminating duplicate data~\cite{xia2016comprehensive,kaur2018data}.Kaur et al.~\cite{kaur2018data} outlined three main types of data deduplication. The first type, text data deduplication, encompasses both file-level and sub-file-level methods. File-level deduplication, also known as single instance storage, removes duplicate files entirely. Sub-file-level deduplication, or block-level deduplication, further divides files into smaller segments and uses hash algorithms to identify and eliminate redundant blocks. 
    The second type, multimedia data deduplication, focuses on removing redundant multimedia content by identifying duplicate images or key frames using visual feature similarity. 
    The third type, cloud-based deduplication, identifies redundant data by comparing new uploads to existing content stored in the cloud. Data deduplication can be implemented through various techniques, such as hash functions, bloom filters and sparse indexing~\cite{xia2016comprehensive}. For instance, Periasamy et al.~\cite{periasamy2021efficient} achieved efficient data deduplication by computing the hash values of content and comparing these values to quickly identify and avoid storing duplicate data. Wei et al.~\cite{wei2010mad2} utilized a Bloom Filter Array to swiftly pre-check data fingerprints, leveraging its efficient capability to distinguish unique from potentially duplicate data, thereby reducing unnecessary accesses to the on-disk Hash Bucket Matrix that serves as the primary data store, and thus accelerating the deduplication process. Lillibridge et al.~\cite{lillibridge2009sparse} presented sparse indexing, which achieves data compression by sampling small chunks within the data stream, indexing these chunks, and using the index information of these chunks to identify and merge larger data segments containing duplicate content. These deduplication techniques can be used in active forgetting, where duplicate data are discarded to optimize storage and retrieval.
    
\end{j_itemize}

In conclusion, active forgetting is effective in improving the storage and retrieval of critical information in memory systems. Active forgetting can be achieved through data compression and deduplication.

\subsection{Parametric Memory}
\label{sec:AI_memory:parametric}

Compared to non-parametric memory stored externally, many AI models form long-term memory by adjusting their parameters to embed the memory within them, known as parametric memory. For example, Hopfield Networks~\cite{hopfield1982neural, ramsauer2020hopfield} build parametric memory by adjusting weights to minimize an energy function, enabling noisy data to be transformed back into its original patterns. Restricted Boltzmann Machines~\cite{hinton2007learning, nair2010rectified, hinton2010practical} create parametric memory by fine-tuning connection weights between the input and hidden layers, reducing the difference between reconstructed and original inputs to capture and replicate key data features. Transformer models~\cite{vaswani2017attention} establish parametric memory by adjusting the weight matrices in the Query, Key, and Value components of the attention mechanism and the feed-forward network, thereby minimizing the loss function and enhancing performance in specific tasks, such as machine translation.

Neural networks excel in complex tasks that involve multi-modal information, particularly in computer vision and natural language processing, and in tasks requiring large-scale training data. Therefore, we focus on neural networks for interpreting parametric memory. Furthermore, recent research on LLMs has provided valuable insights into the handling of parametric memory. Therefore, we draw on LLMs to discuss certain aspects of this topic. The following subsections provide a detailed discussion of the storage, retrieval, and forgetting processes of parametric memory.

\subsubsection{Storage of Parametric Memory}
\label{sec:AI_memory:parametric:storage}

In AI models, particularly in many neural networks, the training process begins by computing the difference between ground-truth labels and model predictions, then updates the parameters using methods like gradient descent with backward propagation; during the inference process, these updated parameters are used to produce model predictions by forward propagation~\cite{lecun2015deep,bengio2017deep,kriegeskorte2019neural}. For instance, in image classification~\cite{lu2007survey}, neural networks update parameters using the images and their associated class labels in the training set. When the model is presented with unseen images, it uses these updated parameters during forward propagation to generate the predicted class for these images. 
In this process, AI models implicitly ``memorize’’ useful information from the training data through parameter updating.
Thus, the type of long-term memory that exists implicitly within AI models is referred to as ``parametric memory''. Parametric memory can implicitly integrate learned information during the inference process, without requiring a mechanism explicitly designed for integrating diverse information, as is necessary in non-parametric memory driven systems~\cite{heinzerlinginui2021language,yu2023generate}.

The storage of parametric memory poses several challenges. 
We discuss these challenges as follows:

\begin{j_itemize}

    \item[$\bullet$] \textbf{Filtering of Inappropriate Data.} Parametric memory, which involves altering a model's parameters, presents significant challenges in erasing relevant information compared to non-parametric memory. Thus, it is essential to filter out inappropriate data before constructing parametric memory. This inappropriate data may include text generated by AI models. Such text may contain more errors than the original training texts and may lead to the production of meaningless outputs or the omission of infrequently mentioned information in the dataset. Recursive use of model-generated texts as training data can precipitate model collapse~\cite{gibney2024ai,shumailov2024ai}. 
    Another concern is poisoned data, which can drastically degrade model performance; for instance, experiments with the IMDB sentiment dataset have demonstrated that introducing just 3 percent poisoned data can escalate the test error from 12 percent to 23 percent~\cite{steinhardt2017certified,cina2023wild}. Lastly, privacy-related concerns arise when training data used for parametric memory can be inadvertently revealed through prompts resembling parts of the training data, leading to potential leaks of sensitive information~\cite{schwarzschild2024rethinking}. Such data may be unsuitable for storage in parametric memory systems, especially in public services~\cite{krishnamurthy2011privacy,pan2020privacy,brown2022does}. To mitigate these risks, employing keyword filtering and discriminative models may effectively prevent the inclusion of such inappropriate data~\cite{androutsopoulos2000experimental,joung2021automated}.

    \item[$\bullet$] \textbf{Storing Long-tail and Time-sensitive Information.} Language models have been shown to efficiently query and ``implicitly'' store factual knowledge akin to knowledge bases~\cite{robertsetal2020much, heinzerlinginui2021language, yu2023generate}.
    The factual knowledge stored in LLMs constitutes their parametric memory.
    However, it is challenging for language models to memorize long-tail information, which refers to the low-frequency data~\cite{kandpal2023large,mallen2023trust}. This occurs in part because such information constitutes a small fraction of the pre-training dataset, making it difficult for models to store and recall it efficiently, even after extensive pre-training.
    Additionally, the static nature of parametric memory can lead to information lag, posing challenges in maintaining the relevance of the stored data as it may not reflect the most current information~\cite{lewis2020retrieval,kasai2024realtime}.
    To address these challenges, researchers have developed methods to mitigate the issues associated with parametric memory.
    Methods like improving the model capacity, increasing the frequency of exposure to examples, and enriching the content of prompt possess the potential to alleviate the difficulties of long-tail information storage and information lag~\cite{carlini2023quantifying}.
    However, it is suggested that storing time-sensitive and accuracy-critical information in non-parametric memory systems might be more effective~\cite{lewis2020retrieval,mallen2023trust,kasai2024realtime}.
    
    \item[$\bullet$] \textbf{Increase of Storage Capacity.} 
    There is a series of papers on \textbf{scaling laws} for the storage efficiency of parametric memory in LLMs (as evidenced by task performance) and the computational resources required during training~\cite{kaplan2020scaling,zhai2022scaling}. 
    Specifically, the information stored in parametric memory is positively correlated with the scale of model parameters $N$, the size of the training dataset $D$, and the amount of computation used during training $C$. Here $C$ required for model training, excluding the embedding process, can be approximated by $C \approx 6\text{NBS}$, where $B$ denotes the batch size, $S$ represents the number of training steps, and 6 signifies the operations involved in forward and backward passes~\cite{kaplan2020scaling}. Initially, as the training computation $C$ increases, the model's performance improves significantly. However, after a certain threshold of $C$ is reached, the rate of performance enhancement begins to decelerate. This relationship can be mathematically formulated as follows:

     {\fontsize{10}{12}\selectfont
    \[ E = a \cdot C^{-b} + c. \]}
    
     Here, $E$ represents the model's error rate that reflects the storage capacity of parametric memory, $a$, $b$ and $c$ are all constants. This formula indicates a diminishing returns issue with parametric memory storage. The improvement in the parametric memory storage capacity brought by the unit computation amount $C$ becomes less and less noticeable as the training progresses. According to Zhai et al.~\cite{zhai2022scaling}, simultaneously scaling up the model parameter size $N$, training data size $D$, and computation $C$ can enhance the storage capability of parametric memory. However, this approach leads to escalating costs, making it increasingly unaffordable, particularly when computational resources are limited. 
     On the other hand, some parameter-efficient fine-tuning~\cite{lialin2023scaling,ding2023parameter} techniques help reduce computational resource usage during incremental training while mitigating the loss of already stored parametric memory~\cite{kalajdzievski2024scaling,ren2024analyzing}. For instance, LoRA~\cite{hu2021lora} adds trainable rank decomposition matrices into Transformer model layers. QLoRA~\cite{dettmers2024qlora} extends LoRA by quantizing the pretrained model to 4-bit precision, further reducing computational resource usage.
     Furthermore, an effective strategy for enhancing long-term memory retention within the constraints of limited computational resources and model size can be realized by strategically distributing non-parametric and parametric memory types. This method involves stratifying the training data by allocating certain types of data, such as long-tail data, to non-parametric memory, while retaining the rest in parametric memory~\cite{izacard2023atlas,mallen2023trust}.

\end{j_itemize}

In summary, parametric memory is stored in AI models' parameters and implicitly integrates learned information. We can improve the reliability of the training set for building parametric memory by filtering out low-quality data, poisoned data, and data that are prone to privacy leakage. 
In comparison with non-parametric memory, parametric memory is not well-suited for storage of long-tail knowledge or time-sensitive information. 
Additionally, the storage capacity of parametric memory can be enhanced by simultaneously increasing the scale of model parameters, the size of the training set, and computational effort.

\subsubsection{Retrieval of Parametric Memory}
\label{sec:AI_memory:parametric:memory_retrieval}

Parametric memory retrieval involves forward propagating queries through AI models to generate predictions, along with several associated challenges: 

\begin{j_itemize}

    \item[$\bullet$] \textbf{Resolving Conflicts in Parametric Memory Retrieval.} 
    Recent research suggests that language models serve a function similar to that of traditional knowledge bases in retrieving information~\cite{petroni2019language,robertsetal2020much,heinzerlinginui2021language,gui2022kat}. For example, if the training data includes ``Bert is a character on Sesame Street'', the model stores this information in its parametric memory. When asked ``Bert is a character on which show?'' the model performs retrieval from its parametric memory to answer, ``Sesame Street''. However, Xu et al.~\cite{xu2024knowledge} observed that retrieval from parametric memory can sometimes conflict with contextual information, such as non-parametric memory fragments or system inputs. For example, if the parametric memory asserts ``Tomatoes are classified solely as fruits'', while either the system input or the retrieved non-parametric memory fragments contend that ``Tomatoes are classified solely as vegetables'', such apparent contradictions
    % \footnote{From Wikipedia: ``Vegetables are parts of plants that are consumed by humans or other animals as food. The original meaning is still commonly used and is applied to plants collectively to refer to all edible plant matter, including the flowers, fruits, stems, leaves, roots, and seeds.'', so a tomato may simultaneously be classified as both a fruit and a vegetable.}
    can lead to the generation of hallucinated information. 
    This issue is extremely difficult to manage, as it involves making judgments about the trustworthiness of various different information sources. 
    Nonetheless, it is possible to address the issue by managing the adherence of language models to conflicting information sources. When prioritizing contextual information, context-aware decoding techniques can be employed to decrease dependence on parametric memory. This approach, detailed by Shi et al.~\cite{shi2023trusting}, modifies the output probability distribution to accentuate differences when the model uses contextual information in generation versus when it does not. This adjustment encourages the model to rely more heavily on the provided context during inference, enhancing its accuracy and relevance. Conversely, if adherence is based on factual consistency, language models can be equipped with self-discrimination to evaluate and reconcile parametric and context information~\cite{pan2023attacking,hong2023so}.

    \item[$\bullet$] \textbf{Resolving Hallucinations in Parametric Memory Retrieval.} In addition to conflicts of parametric memory, hallucinations can also be caused by low-quality training data and poor model generalization abilities~\cite{zhang2023siren}. According to Zhang et al.~\cite{zhang2023siren}, hallucinations refer to the phenomenon where the content retrieved and generated through parametric memory does not align with the source input provided by the user~\cite{adlakhaetal2024evaluating}, previously generated content~\cite{liuetal2022token}, or established facts~\cite{minetal2023factscore}. Mitigating hallucinations can be achieved through a variety of strategies~\cite{zhang2023siren}. One strategy is ensuring the factuality in the pre-training corpus~\cite{gardent2017creating,parikhetal2020totto,lee2022factuality}. Another approach is constructing a reward model based on human feedback, and using reinforcement learning algorithms with a reward model that encourages the target model to acknowledge its uncertainty, especially in scenarios where correctness cannot be guaranteed~\cite{schulman2017proximal,fernandes2023bridging}. 
    The previously mentioned methods for resolving conflicts can also be used to mitigate hallucinations. 
    These mitigation methods are quite similar to the recognition stage of the Generation-Recognition Theory~\cite{shiffrin1969memory,underwood1969memory,tulving1973encoding} described in Sec.~\ref{sec:human_memory:processing:retrieval}, which judges the correctness of retrieved content in human memory retrieval.

    \item[$\bullet$] \textbf{Enhancement of Query for Parametric Memory Retrieval.} Tulving et al.~\cite{tulving1973encoding,bower1972encoding} highlight the importance of cues for memory retrieval in the human brain. Similarly, effective queries enhance AI model's parametric memory retrieval, benefiting task performance. In image classification~\cite{lu2007survey}, the query is the image itself, so image denoising~\cite{tian2020deep} and quality enhancement~\cite{sharma2018classification} improve the accuracy of parametric memory retrieval in this task. 
    In content generation~\cite{kenton2019bert,yu2023generate,wen2024hard}, the query for parametric memory retrieval is a prompt, which is a specific input or instruction given to an AI model to guide its output and serves as the starting point or the initial information that the model uses to generate the desired content.
    Therefore, prompt engineering~\cite{pryzant2023automatic,wen2024hard} can be adopted to refine these inputs to achieve better performance and higher quality results.

\end{j_itemize}

In summary, parametric memory retrieval can be achieved through forward propagation (within neural networks). 
We discuss three challenges in leveraging parametric memory: conflicts, hallucinations, and query enhancements.
We review existing works to offer insights into the potential solutions to these challenges.

\subsubsection{Forgetting of Parametric Memory}
\label{sec:AI_memory:parametric:memory_forgetting}

Unlike non-parametric memory, precisely managing the forgetting process of parametric memory through specific mechanisms is challenging. In multi-task learning with neural networks, updating parameters for new tasks may introduce errors into the parametric memory of previously learned tasks, potentially leading to catastrophic forgetting. Aleixo et al.~\cite{aleixo2023catastrophic} summarized four primary methods to mitigate catastrophic forgetting: rehearsal, distance-based methods, sub-networks, and dynamic networks. In addition to these methods, our findings indicate that curriculum learning can alleviate catastrophic forgetting. The following paragraphs explore these methods.

\begin{figure*}
\centering
\includegraphics[scale=0.64]{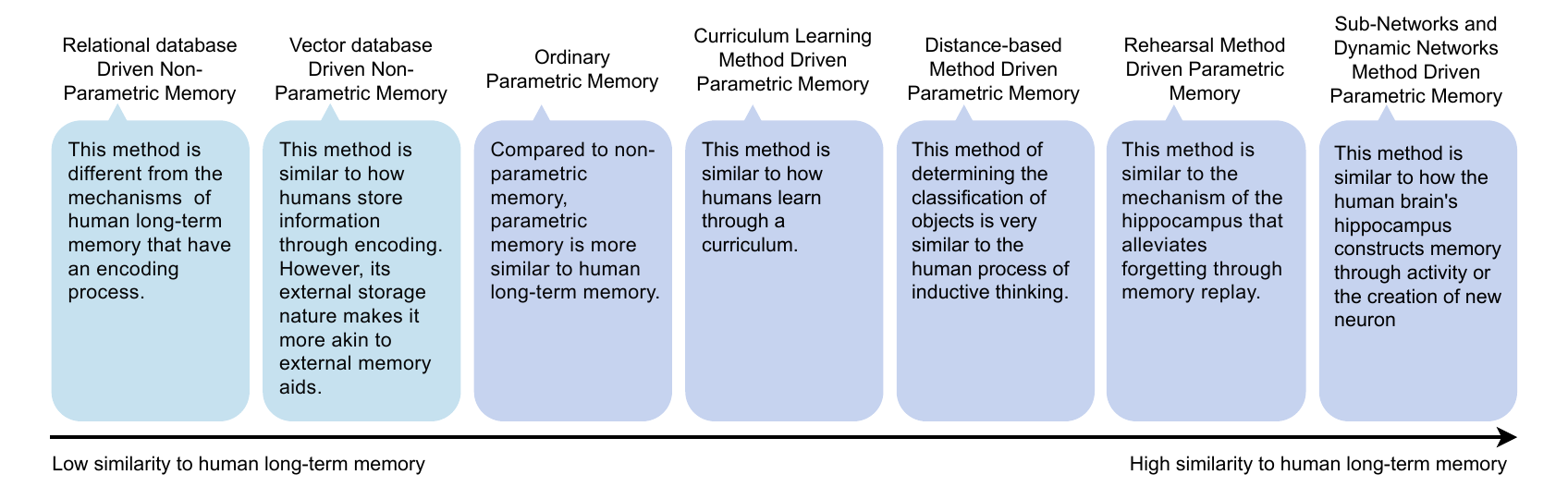}
\caption{The similarities between different types of AI long-term memory and human long-term memory. Types of Non-Parametric Memory refer to Sec.~\ref{sec:AI_memory:non_parametric:memory_storage}, and types of Parametric Memory refer to Sec.~\ref{sec:AI_memory:parametric:memory_forgetting}.}
\label{fig:similarity}
\end{figure*}

\begin{j_itemize}

    \item[$\bullet$] \textbf{Rehearsal Method.} The rehearsal method, based on Aleixo et al.'s theory~\cite{aleixo2023catastrophic}, involves obtaining representative samples of previous tasks for incremental training to mitigate catastrophic forgetting. These samples can either be sourced directly from the original data~\cite{hinton1987using,lopez2017gradient,nguyen2018variational,de2019episodic,prabhu2020gdumb,roscow2021learning} or generated synthetically~\cite{mellado2017pseudorehearsal,zhu2021prototype,roscow2021learning,spens2024generative}. These samples are used in incremental training to maintain knowledge learned in the previous training process when training on new tasks. For instance, Hinton et al.~\cite{hinton1987using} used fast weights to quickly deblur old memory stored in slow weights that have been blurred by the learning of new samples, based on a small subset of the old samples. Spens et al.~\cite{spens2024generative} developed an autoassociative network that can encode sensory input and replay related episodic memory using a modern Hopfield network~\cite{ramsauer2020hopfield}. This autoassociative network assists in training the generative models through the process of replay. Roscow et al.~\cite{roscow2021learning} stored the interaction data comprising states, actions, time steps, and rewards in a memory buffer, and select samples from it to prevent catastrophic forgetting, either randomly or based on criteria such as temporal differences or error rates. They also discussed the use of generative models to produce data that match the parameter distribution of historical input data. This method can reduce the need for storage space when replaying training data from old tasks to the target model, as it does not require storing exact copies of training samples. The rehearsal method is similar to the mechanism of the hippocampus that alleviates forgetting through memory replay~\cite{ji2007coordinated,olafsdottir2018role}.

    \item[$\bullet$] \textbf{Distance-based Method.} The distance-based method, based on Aleixo et al.'s theory~\cite{aleixo2023catastrophic}, is suitable for multi-class continual learning tasks. Its objective is to minimize the distance between data points within the same class and to maximize the distance between data points of different classes, thereby maintaining the distinctiveness between classes and avoiding catastrophic forgetting of class attribution information during continual learning. For example, both Cheraghian et al.~\cite{cheraghian2021semantic} and Ni et al.~\cite{ni2024enhancing} used fixed semantic embeddings to represent class information, which ensured a degree of separability between data from different classes. Their approaches helped alleviate representation drift and mitigate catastrophic forgetting. Shi et al.~\cite{shi2022mimicking} minimizes redundant information between features by reducing the correlation between data representations within the same class, thereby encouraging a more independent distribution of data points in the feature space. As a result, representations of the same class exhibit tighter clustering due to diminished correlation, while representations of different classes are more dispersed, occupying distinct regions of the feature space. This enhances the overall discriminability between classes. Song et al.~\cite{song2023learning} expanded the base classes using predefined transformations, such as rotations and color permutations, to create virtual classes. They provided data from these expanded classes for the model to learn class discrimination. This approach helped the newly learned classes reduce overlap with the feature space of existing classes, thereby mitigating the risk of catastrophic forgetting. The distance-based method resembles how humans construct long-term memory by building associations between pieces of information~\cite{shiffrin1969memory,tulving1972episodic}.

    \item[$\bullet$] \textbf{Sub-Networks Method.} The sub-networks method, based on Aleixo et al.'s theory~\cite{aleixo2023catastrophic}, mitigates catastrophic forgetting by leveraging the idea that different tasks rely on distinct sets of weight parameters within the same model. For example, task $A$'s accuracy may be highly dependent on weight parameter $a$ but less so on parameter $b$, while task $B$'s accuracy may be highly dependent on parameter $b$ but less so on parameter $a$. Therefore, during training for task $A$, we should focus on updating parameter $a$ and avoid changing parameter $b$. Similarly, during training for task $B$, we should focus on updating parameter $b$ and avoid changing parameter $a$. This approach of designating specific weight parameters for different tasks is known as the Sub-Networks Method. 
    One method to implement sub-networks is through normalization techniques. Pham et al.~\cite{pham2022continual} apply group normalization to feature channel groups prior to batch normalization, helping to balance knowledge transfer. It enhances adaptation to new tasks with significantly different data distributions while preserving the model’s performance on previously learned tasks. Another method involves Orthogonal Weights Modification. Zeng et al.~\cite{zeng2019continual} and Shen et al.~\cite{shen2021generative} construct a matrix representing all previously trained input vectors when training on a new task. By utilizing an orthogonal projector, weight updates are confined to directions orthogonal to this matrix, thus preventing negative interference with earlier tasks. A third method employs adaptive parameters to control changes in model weights. Adel et al.~\cite{Adel2020Continual} introduced an adaptive binary variable that determines whether the parameters of each neuron in a deep network should update during the learning of a new task. Xue et al.~\cite{xue2022meta} addressed catastrophic forgetting by dynamically generating adaptive attention masks for each new task, targeting key parameters in the Vision Transformer (ViT). Some studies indicated that Low-Rank Adaptation (LoRA) alleviates catastrophic forgetting in incremental training specifically for Transformer models by integrating tunable adapter modules into certain layers, such as attention layers and feed-forward neural network (FFN) layers, while minimally affecting the original model weights~\cite{hu2021lora,kalajdzievski2024scaling,ren2024analyzing}. The concept of sub-networks is analogous to the activity in specific regions of the hippocampus in the human brain, which is involved in the formation of new memory~\cite{lavenex2000hippocampal,rolls2006spatial, meira2018hippocampal}.

    \item[$\bullet$] \textbf{Dynamic Networks Method.} The dynamic networks method, discussed by Aleixo et al.~\cite{aleixo2023catastrophic}, expands the parameters when learning new tasks to mitigate catastrophic forgetting. For instance, Roy et al. proposed Tree-CNN for handling the introduction of new classes~\cite{roy2020tree}. When a new class is introduced and has a low association with existing child nodes, or the current child nodes are full, the Tree-CNN expands its parameters by adding the new class as a new child node to the original structure. Hu et al.~\cite{hu2023dense} capture task-specific features by utilizing expert networks, ensuring that networks associated with earlier tasks remain unaltered. During inference, they interconnect the expert networks, enabling the exchange of feature information across tasks. Ye et al.~\cite{ye2023self} utilized the diversity of knowledge among experts to dynamically expand the expert network. Specifically, they create new experts based on the current data stream. If the minimum similarity between the new expert and existing experts, based on the output of the current data stream, is greater than or equal to a preset threshold, it indicates sufficient knowledge diversity, and the new expert is added to the ensemble. In addition to dynamically expanding experts based on new tasks, the expert network can also grow by learning task-specific routing for the experts. Yu et al.~\cite{yu2024boosting} proposed a method that integrates Mixture-of-Experts (MoE) adapters into a pre-trained CLIP model. These adapters function as experts within the model and can adapt to new tasks using an incremental activate-freeze strategy. In this approach, when new tasks are introduced, the model selectively activates or freezes experts based on the routers' output distribution. This allows the model to dynamically expand and fine-tune itself to adapt to the needs of new tasks while preserving knowledge of previous tasks. The dynamic networks method mirrors the way the hippocampus in the human brain stores new memory through the formation of new neurons~\cite{deisseroth2004excitation}.

    \item[$\bullet$] \textbf{Curriculum Learning Method.} Curriculum Learning mimics the learning order in human education by training the model in a specific data sequence, which enhances the model's adaptability to new tasks and improves its resistance to catastrophic forgetting~\cite{parisi2019continual,wang2021survey}. For example, Lisicki et al.~\cite{lisicki2020evaluating} employs an adaptive staircase strategy for curriculum learning, gradually increasing problem difficulty and periodically reviewing simpler tasks during training to balance the acquisition of new knowledge with the retention of old knowledge, thereby alleviating catastrophic forgetting. Bhat et al.~\cite{bhat2021cilea} leveraged features extracted by Convolutional Neural Networks (CNNs)~\cite{li2021survey} to measure class similarity and arranged the learning order from the most to the least similar, thereby linking new knowledge with prior knowledge to mitigate catastrophic forgetting. Saglietti et al.~\cite{saglietti2022analytical} drawing on the concept of synaptic consolidation from biology, apply L2 regularization during curriculum phase transitions to guide the current weights towards the direction of the weights learned in previous stages, thereby facilitating the retention of memory for previously learned content in new learning phases.

\end{j_itemize}

In summary, the tendency of parametric memory to experience catastrophic forgetting is similar to the human brain's susceptibility to memory forgetting due to interference~\cite{underwood1957interference}. According to Aleixo et al.~\cite{aleixo2023catastrophic}, the mitigation of catastrophic forgetting in parametric memory can be achieved through incremental training of previous tasks with rehearsal samples, maintaining class separability, partitioning task-specific parameter subsets, and dynamically expanding the models based on the learning needs of new tasks. Additionally, Curriculum Learning can also be utilized to alleviate catastrophic forgetting.

\subsection{Summary}

AI long-term memory can be categorized into non-parametric and parametric memory based on their storage forms. Non-parametric memory is stored externally in mediums such as databases, independent of the AI model, whereas parametric memory is stored as parameters within the AI model itself. These two AI long-term memory processing methods are shown in Fig.~\ref{fig:foundation}, and reflected in the non-parametric memory and parametric memory processing sections of Fig.~\ref{fig:lit_surv}. While both non-parametric and parametric memory align with the characteristic of human long-term memory in ``storing information relatively permanently''~\cite{atkinson1968human}, the encoding-specific nature of human memory~\cite{tulving1973encoding} leads to different degrees of similarity between human long-term memory and these AI long-term memory. 

We organize these similarities in Fig.~\ref{fig:similarity}. Sub network driven parametric memory and dynamic network driven parametric memory exhibit the highest similarity to human long-term memory due to their highly encoded nature. Relational database driven non-parametric memory exhibits the least similarity to human long-term memory because it is stored independently of the AI model in a relatively structured format, which differs significantly from the encoded storage of human long-term memory. Instead, it functions more like an external memory aid for human memory~\cite{intons1992external,intons2014external}.

Furthermore, compared to non-parametric memory, the characteristics of parametric memory storing information implicitly within AI models are more akin to how the human brain produces new neurons or undergoes chemical changes~\cite{lavenex2000hippocampal,deisseroth2004excitation,rolls2006spatial,meira2018hippocampal}.

\section{Long-term Memory of AI: on Human Perspectives}
\label{sec:hierarchy_ai_memory}

\begin{figure*}
\centering
\includegraphics[scale=0.64]{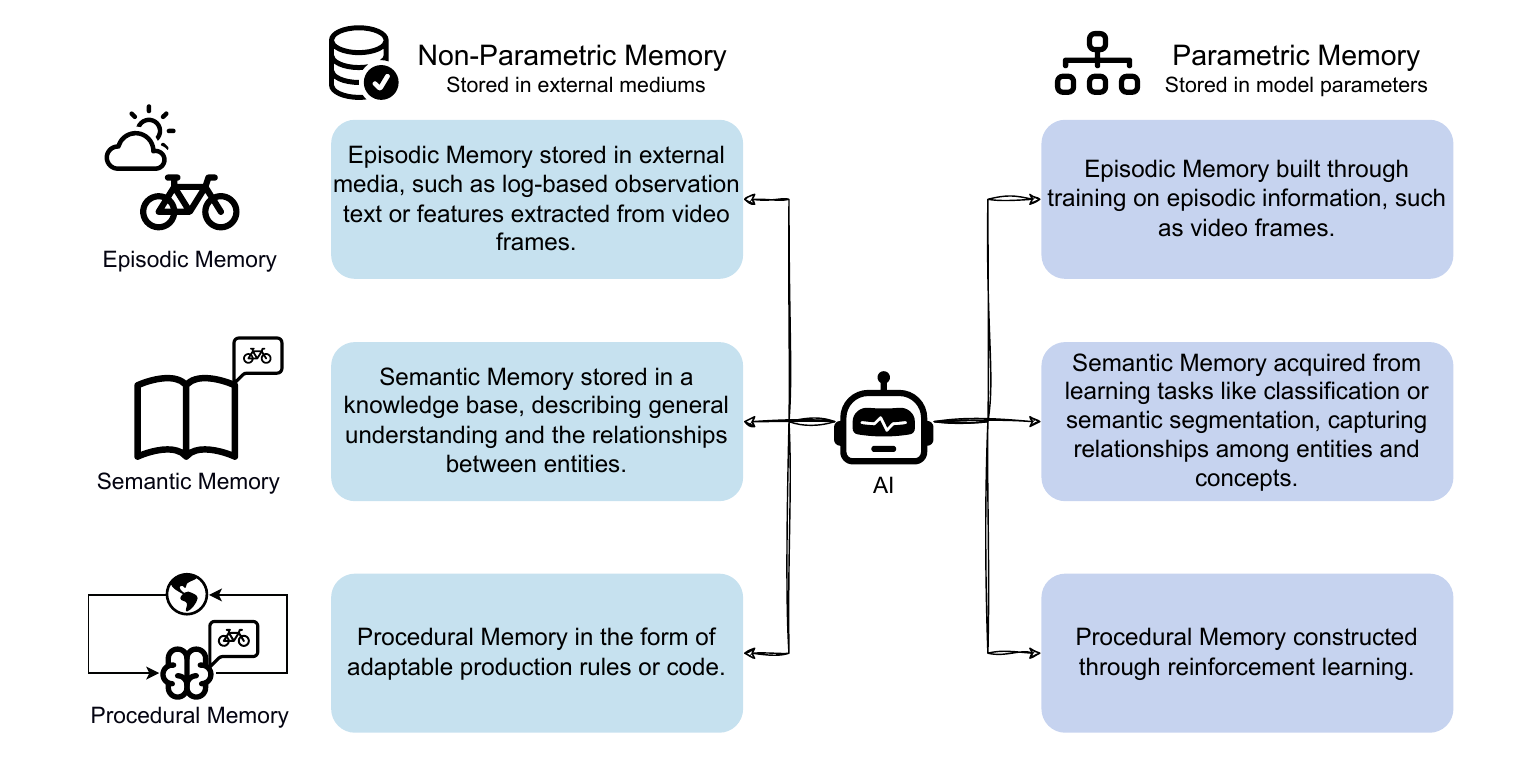}
\caption{Taxonomy of AI parametric memory and non-parametric memory constructed based on three key types of human long-term memory: episodic memory (Sec.~\ref{sec:hierarchy_ai_memory:episodic}), semantic memory (Sec.~\ref{sec:hierarchy_ai_memory:semantic}), and procedural memory (Sec.~\ref{sec:hierarchy_ai_memory:procedural}).}
\label{fig:ai_episodic_semantic_procedural_memory}
\end{figure*}

AI long-term memory can be classified both by storage format and by analogy to human long-term memory types. Based on storage formats, it is divided into non-parametric memory (Sec.~\ref{sec:AI_memory:non_parametric}) and parametric memory (Sec.~\ref{sec:AI_memory:parametric}). In terms of human memory analogies, AI memory reflects key components of human long-term memory: episodic, semantic, and procedural memory (Sec.~\ref{sec:human_memory:hierarchy:long_term_memory}). Episodic memory in AI systems supports learning from past experiences (Sec.~\ref{sec:hierarchy_ai_memory:episodic}), semantic memory enables the development of generalized knowledge (Sec.~\ref{sec:hierarchy_ai_memory:semantic}), and procedural memory facilitates learning through feedback mechanisms (Sec.~\ref{sec:hierarchy_ai_memory:procedural}).

In the following subsections, we explain in detail how AI long-term memory corresponds to human episodic, semantic, and procedural memory. 
We demonstrate these mappings in the Fig.~\ref{fig:ai_episodic_semantic_procedural_memory} and Category sections of Fig.~\ref{fig:lit_surv}.
By establishing these connections, we aim to map the relationship between AI and human long-term memory systems.

\subsection{Episodic Memory}
\label{sec:hierarchy_ai_memory:episodic}

Similar to human brains, episodic memory helps AI record past experiences, events, and contexts~\cite{tulving1972episodic,andrew2007episodic}. Episodic memory of AI has both non-parametric and parametric format.

\begin{j_itemize}

    \item[$\bullet$] \textbf{Parametric Episodic Memory.} Because the development of parametric memory hinges on parameter updates during training, and the goal of such training is generally to acquire general knowledge rather than to gain subjective experiences, there is a scarcity of research on parametric episodic memory~\cite{li2021survey,nandwani2021review,guo2018review,li2020survey}. For instance, AI models are typically trained to identify a ``cat''~\cite{he2016deep,li2021survey}, apply the characteristics of a ``cat'' to solve reasoning challenges~\cite{heinzerlinginui2021language,mallen2023trust}, or execute actions associated with ``cats''~\cite{kaelbling1996reinforcement}. In contrast, they are less commonly trained to recall a specific past event involving a ``cat'' at a particular time and location. Nevertheless, certain studies can be considered to have employed parametric episodic memory. For example, Sun et al.~\cite{sun2024learning} proposed a recurrent neural network that updates its model parameters during the inference phase, with the potential to ``memorize'' episodic information. When processing temporal data inputs, such as video frames, the model parameters are adjusted. This type of model could potentially be used to construct episodic memory in a parametric format. Di et al.~\cite{di2024grounded} utilized automatically generated question-answer pairs with timestamps as training data to build the episodic memory capability of a long egocentric video question answering system, where the episodic memory involved belongs to parametric memory. Spens et al.~\cite{spens2024generative} proposed a computational model for memory construction and consolidation that uses a modern Hopfield network (MHN)~\cite{ramsauer2020hopfield} to encode episodic memory into its parameters. This MHN functions as an autoassociative network, trained to reconstruct episodic memory by capturing the statistical structure of experienced events.
    
    \item[$\bullet$] \textbf{Non-Parametric Episodic Memory.} In contrast, non-parametric episodic memory is widely employed in AI systems. It involves utilizing event-specific data to support task execution. This memory type offers experiential knowledge, functioning as an extension to the inherently limited context of AI systems~\cite{andrew2007episodic,das2024larimar,fountas2024human}, which can enhance the system's capabilities across a range of applications. 
    For instance, in visual understanding tasks, AI systems can access visual features from previous frames that are relevant to the current scene. This allows for improved comprehension and more accurate responses regarding past events~\cite{mezghan2022memory,cheng2022xmem,song2023moviechat,shen2023encode,jilan2024egoInstructor,he2024ma}. 
    Furthermore, non-parametric episodic memory plays a pivotal role in shaping AI systems' behavioral strategies~\cite{vere1990basic,ho2003comparing,andrew2007episodic,roscow2021learning,shinn2024reflexion}. It enables AI to make decisions informed by experiences, aligning their actions with expected outcomes. For example, Andrew et al.~\cite{andrew2007episodic} proposed that in a tank game, an agent can use episodic memory to predict the environment that might be observed when radar is activated, and utilize this information to optimize the use of radar, thereby reducing energy consumption. Additionally, non-parametric episodic memory can serve as rehearsal samples to alleviate catastrophic forgetting in parametric memory~\cite{lopez2017gradient,nguyen2018variational,de2019episodic}. For instance, Nguyen et al.~\cite{nguyen2018variational} employed episodic memory as a mechanism for storing key data points from previous tasks, allowing the model to periodically revisit these data points when learning new tasks, thus ensuring the retention of knowledge from old tasks.

\end{j_itemize}

In summary, episodic memory in AI models plays a crucial role in capturing and storing past events and experiences. Compared to parametric memory, episodic memory more commonly exists in the form of non-parametric memory. It can be used to extend the context of AI systems, shape AI systems' behavioral strategies, and serve as training samples to prevent catastrophic forgetting.

\subsection{Semantic Memory}
\label{sec:hierarchy_ai_memory:semantic}

Unlike episodic memory, which records personal experiences, semantic memory helps AI build and understand general knowledge~\cite{tulving1972episodic,wang2016semantic}.

\begin{j_itemize}

    \item[$\bullet$] \textbf{Non-Parametric Semantic Memory.} External knowledge from sources such as knowledge bases allows AI systems to improve their performance on tasks that require objective information, such as content generation~\cite{trisedya2020sentence,wenhu2022reimage,yu2022survey,rongjie2023makeanaudio} and question answering~\cite{qa2020karpukhin,xixin2022gmtkbqa,yang2023care,luo2023chatkbqa}. This form of external knowledge is often termed non-parametric semantic memory, as it aids AI in understanding objective information, similar to the role of semantic memory in humans. Furthermore, when structured in graph format, non-parametric semantic memory offers greater efficiency in both storage and retrieval~\cite{edge2024local,gutierrez2024hipporag}.

    \item[$\bullet$] \textbf{Parametric Semantic Memory.} As introduced in Sec.~\ref{sec:human_memory:hierarchy:long_term_memory}, human semantic memory involves constructing ``relationship'' between pieces of information~\cite{shiffrin1969memory,tulving1972episodic}. AI models build their understanding of these ``relationship'' by being trained for tasks such as classification, segmentation, and content generation. Thus, training these models on such tasks is a process of constructing semantic memory. Classification tasks enable AI models to recognize the ``relationship'' between specific instances and abstract concepts~\cite{he2016deep,li2021survey,nandwani2021review}. For instance, a well-trained animal image classification model can determine whether an unseen image belongs to the ``cat'' category. Similarly, semantic segmentation tasks allow AI models to extract subsets of information related to abstract concepts from complete information~\cite{long2015fully,guo2018review}. For example, a well-trained semantic segmentation model for pedestrian detection can identify specific regions in urban traffic images that correspond to the ``pedestrian'' category.
    During the training process, model parameters are regularly updated, reinforcing the model's ability to understand ``relationship'' between pieces of information. This process corresponds to the construction of parametric semantic memory.
    
\end{j_itemize}

Numerous studies on language models highlighted the distinctions between non-parametric and parametric semantic memory, enhancing our understanding of the semantic memory in AI models~\cite{heinzerlinginui2021language,yu2023generate,mallen2023trust,kasai2024realtime}. 
When a language model is queried about the location of Australia, it can utilize non-parametric semantic memory. For example, it might retrieve the fact ``Australia is located in the Southern Hemisphere'' from an external knowledge base to generate the answer ``Southern Hemisphere''. 
Alternatively, if this information is included in the training data, the language model can directly generate the answer ``Southern Hemisphere'' when queried about Australia's location, demonstrating the use of parametric memory. 
Parametric semantic memory typically exhibits superior generalization capabilities, enabling the model to infer answers for unseen queries based on foundational knowledge. 
For instance, if the language model learns through parametric memory that ``whales give birth to live young'' and ``most mammals give birth to live young'', it will likely infer that ``whales are mammals'' when queried. 
In contrast, non-parametric semantic memory excels at storing long-tail knowledge and offers greater extensibility and timeliness. For example, a database of non-parametric semantic memory can be maintained for the film industry. 
This type of database can be regularly updated with introductions of new movies and information about lesser-known films, which are time-sensitive or long-tail information, through simple data operations, thereby eliminating the need to frequently retrain the model with new data.

Semantic and episodic memory in AI models can be interconverted through specific methods. Huang et al.~\cite{rongjie2023makeanaudio} introduced an algorithm named Make-an-Audio, which uses scenario-descriptive prompts as input and a diffusion model to generate high-definition audio. In Make-an-Audio, textual information is transformed into audio that reflects situational content, illustrating how episodic memory can be constructed from semantic memory. Conversely, Wang et al.~\cite{wang2016semantic} constructed an expandable semantic memory network. They traversed episodic memory to find descriptions of instances associated with the original semantic memory network, thereby expanding the semantic memory network. This process converts episodic memory into semantic memory.

In summary, semantic memory aids AI models in managing general understanding and comprehending ``relationship'' of information. Non-parametric semantic memory serves as an external reservoir of knowledge for AI systems, offering greater extensibility and timeliness, and parametric semantic memory enables AI models to delve deeper into the understanding of ``relationship'' between pieces of information, exhibiting superior generalization capabilities. Moreover, semantic memory and episodic memory of AI can be interconverted to a certain extent.

\begin{table*}

    \centering
    \scalebox{0.68}{
    \begin{tabular}{lrrrrrrrrrr}
        \toprule
              & ACT~\cite{anderson2013architecture} & ACT-R~\cite{anderson1997act} & Soar~\cite{laird2008extending} & Sigma~\cite{rosenbloom2016sigma} & SMoM~\cite{laird2017standard} & Episodic Soar~\cite{andrew2007episodic} & Graphical Soar~\cite{rosenbloom2011memory} & RL Sigma~\cite{pynadath2014reinforcement} & CoALA~\cite{sumers2023cognitive} & \textbf{SALM} (Ours) \\ 
             \midrule
            Episodic Memory (NP) & \cmark & \cmark & \xmark & \cmark & \cmark & \cmark & \cmark & \xmark & \cmark & \cmark \\
            Semantic Memory (NP) & \cmark & \cmark & \xmark & \cmark & \cmark & \xmark & \cmark & \xmark & \cmark & \cmark \\
            Procedural Memory (NP) & \cmark & \cmark & \cmark & \cmark & \cmark & \xmark & \cmark & \xmark & \cmark & \cmark \\
            Episodic Memory (P) & \xmark & \xmark & \xmark & \xmark & \xmark & \xmark & \xmark & \xmark & \xmark & \cmark \\
            Semantic Memory (P) & \xmark & \xmark & \xmark & \cmark & \xmark & \xmark & \xmark & \xmark & \xmark & \cmark \\
            Procedural Memory (P) & \xmark & \xmark & \xmark & \xmark & \cmark & \xmark & \xmark & \cmark & \cmark & \cmark \\
            Storage & \cmark & \cmark & \cmark & \cmark & \cmark & \cmark & \cmark & \cmark & \cmark & \cmark \\
            Retrieval & \cmark & \cmark & \cmark & \cmark & \cmark & \cmark & \cmark & \cmark & \cmark & \cmark \\
            Forgetting & \xmark & \xmark & \xmark & \xmark & \xmark & \xmark & \xmark & \xmark & \xmark & \cmark \\
            Systematic Adaptability & \xmark & \xmark & \xmark & \xmark & \xmark & \xmark & \xmark & \xmark & \xmark & \cmark \\
        \bottomrule
    \end{tabular}
    }
    \caption{Modules related to AI long-term memory in some cognitive architectures work. \textbf{(NP)} stands for Non-Parametric Memory, and \textbf{(P)} stands for Parametric Memory. \textbf{SALM} stands for Cognitive Architecture of \textbf{S}elf-\textbf{A}daptive \textbf{L}ong-term \textbf{M}emory proposed by this work. A check mark indicates presence and a cross mark indicates absence.}
    \label{tab:cognitive_arch}
    
\end{table*}

\subsection{Procedural Memory}
\label{sec:hierarchy_ai_memory:procedural}

Human procedural memory refers to the acquisition of motor skills~\cite{tulving1985memory}, which requires feedback for its development. For example, mastering balance is crucial when learning to a ride bicycle. During the learning process, feedback comes from the sense of balance while riding. Humans seek positive feedback from achieving good balance by adjusting their riding actions, thereby forming the procedural memory for riding a bicycle. Procedural memory of AI can also be constructed based on similar feedback mechanisms.

\begin{j_itemize}

    \item[$\bullet$] \textbf{Parametric Procedural Memory.} The loop of ``primitive action - feedback - update procedural memory - improved action'' described in the bicycle riding example is similar to reinforcement learning in AI~\cite{kaelbling1996reinforcement,liu2021automated,liu2022feature}. In reinforcement learning, positive feedback from a certain action results in a positive reward. Based on this positive reward, the parameters of the policy model are adjusted to increase the probability of selecting this action in the corresponding state. Conversely, if it is negative feedback, the parameters of the model will be adjusted to decrease the probability of selecting this action. The parameter adjustments involved in this process constitute the construction of parametric procedural memory.

    \item[$\bullet$] \textbf{Non-Parametric Procedural Memory.} Some non-parametric components can also be employed to construct procedural memory, such as production rules~\cite{davis1977production,quinlan1987generating,ceri1991deriving,laird2008extending}. A production rule is symbolically represented by a condition and an associated action. When the condition is satisfied, the corresponding action will be executed. This mechanism reflects how human procedural memory operates by responding to specific environmental cues, as discussed in the preceding paragraphs. Additionally, some studies revealed that agents can better adapt to their environment by continuously adjusting their code, a process regarded as leveraging non-parametric components to develop procedural memory~\cite{wang2023voyager,sumers2023cognitive}.

\end{j_itemize}

Procedural memory and episodic memory in AI systems can be interchangeable to some extent. Some studies highlighted the functional similarities between non-parametric episodic memory, represented by experiential data, and parametric procedural memory, represented by updated parameters in reinforcement learning~\cite{sumers2023cognitive, shinn2024reflexion}. Shinn et al.~\cite{shinn2024reflexion} utilized experiential texts stored in an episodic memory buffer to guide the actions of the agent based on language models. The agent can reflect on environmental feedback and update the experiential texts to improve its action strategy over time. For example, if the agent is tasked with finding an item in drawers, it may first search drawer 1 without success and record the note ``drawer 1 has no item'' in experiential texts. When the agent finds the item in drawer 3, it updates its experiential texts to ``drawer 3 has the item'' after reflection. In future tasks, the agent will refer to its experiential text and directly open drawer 3. This iterative process is similar to how reinforcement learning optimizes action strategies based on feedback. 

Other studies showed that parametric procedural memory can be enhanced by episodic and semantic memory. Wang et al.~\cite{wang2016semantic} found that semantic memory can provide relevant background information for reinforcement learning. This helps in the development of parametric procedural memory. For example, in first-person shooter games, semantic memory might include information about weapon ranges, which aids in developing weapon selection strategies in reinforcement learning. Savinov et al.~\cite{savinov2019episodic} used episodic memory to help build procedural memory. Their approach involves reinforcement learning to help agents explore new pathways in their environment. During this process, the agent compares new observations with those stored in episodic memory to evaluate the novelty of new observations. The more novel the observations are, the higher the reward scores they receive. The reward scores are then used in reinforcement learning. This method encourages agents to explore new pathways by constructing specific parametric procedural memory with the aid of episodic memory. Roscow et al.~\cite{roscow2021learning} discussed the use of experience replay within Deep Q Network, a category of reinforcement learning algorithms. They meticulously documented the agent's states, actions, time steps, and reward signals into a memory buffer. During the learning phase, they drew samples in mini-batches from this buffer to augment the DQN's performance. This approach aligns with the enhancement of parametric procedural memory (embodied in the DQN's parameters) through the integration of non-parametric episodic memory (as represented by the experience data stored in the buffer).

In summary, procedural memory enables AI systems to select appropriate actions through feedback. Parametric procedural memory can be constructed through reinforcement learning, while production rules and code can be utilized to build non-parametric procedural memory. 
For AI systems, procedural memory shares functional similarities with episodic memory to some extent and can benefit from episodic and semantic memory.

\subsection{Summary}

Human long-term memory is a multifaceted construct, encompassing episodic memory for recording personal experiences, semantic memory for storing general information, and procedural memory linked to motor skills. The AI long-term memory research corresponds well with episodic memory (Sec.~\ref{sec:hierarchy_ai_memory:episodic}), semantic memory (Sec.~\ref{sec:hierarchy_ai_memory:semantic}), and procedural memory (Sec.~\ref{sec:hierarchy_ai_memory:procedural}). Furthermore, the episodic, semantic, and procedural memory of AI are enriched by both non-parametric and parametric memory, showcasing a more complex and versatile long-term memory architecture. This variety within AI long-term memory provides a broad spectrum of choices for the development of related systems. 

% \section{AI Long-term Memory in Cognitive Architecture}
% \section{Long-term Memory in Cognitive Architectures}
\section{A New Cognitive Architecture for Long-term Memory}
\label{sec:hierarchy_ai_memory:cognitive_arc}

Cognitive architectures~\cite{samsonovich2010toward,laird2017standard,kotseruba202040} provide a foundation for building AI systems by incorporating insights from cognitive science, neuroscience, and artificial intelligence. Unlike traditional task-oriented designs, cognitive architectures use human cognitive modules as models to design and describe AI systems, following a bottom-up framework where modules like long-term memory work together to complete tasks. 

Cognitive architectures provide diverse mechanisms for long-term memory processing. For example, Adaptive Control of Thought (ACT) is an early cognitive architecture enabling advanced cognitive tasks like imagery and deduction through a unified system of long-term memory, including both declarative and procedural types~\cite{anderson2013architecture}. Its enhanced version, ACT-R, describes how information is retrieved from long-term memory through ``activation computation'' and ``conflict resolution''~\cite{lebiere1993connectionist,anderson1997act,whitehill2013understanding}. This process activates relevant knowledge, and conflict resolution determines the most reliable information when discrepancies arise. Soar and Sigma are other architectures that incorporate various memory types, with Sigma using factor graphs for its semantic memory~\cite{rosenbloom2016sigma}. 
Standard Model of the Mind (SMoM) consists of independent long-term memory modules, including declarative and procedural memory, which interact with working memory to enhance storage and retrieval~\cite{laird2017standard}. CoALA, which is built on large language models (LLMs), enhances episodic and semantic memory by leveraging external data~\cite{sumers2023cognitive}. However, these cognitive architectures have certain limitations. 
As summarized in Table~\ref{tab:cognitive_arch}, they do not encompass all types of long-term memory and processing mechanisms. Additionally, they lack a systematic adaptive mechanism for long-term memory processing, including strategies that enable the system to determine processing operations, such as selecting the storage format, choosing retrieval sources, and identifying content for forgetting.

% In response to these limitations, we propose the Cognitive Architecture of \textbf{S}elf-\textbf{A}daptive \textbf{L}ong-term \textbf{M}emory (\textbf{SALM}). This framework introduces two key improvements. First, it clarifies the processing mechanisms of storage, retrieval, and forgetting across three key types of long-term memory—episodic, semantic, and procedural—while addressing both parametric and non-parametric memory forms. Second, it features an adaptive module that optimizes these processing mechanisms based on environmental feedback, showcasing a capacity for adaptive long-term memory that potentially surpasses human memory adaptability. As shown in Table~\ref{tab:cognitive_arch}, \textbf{SALM} offers a promising theoretical basis for the next generation of AI systems with self-adaptive long-term memory.

\begin{figure}
\centering
\includegraphics[scale=0.65]{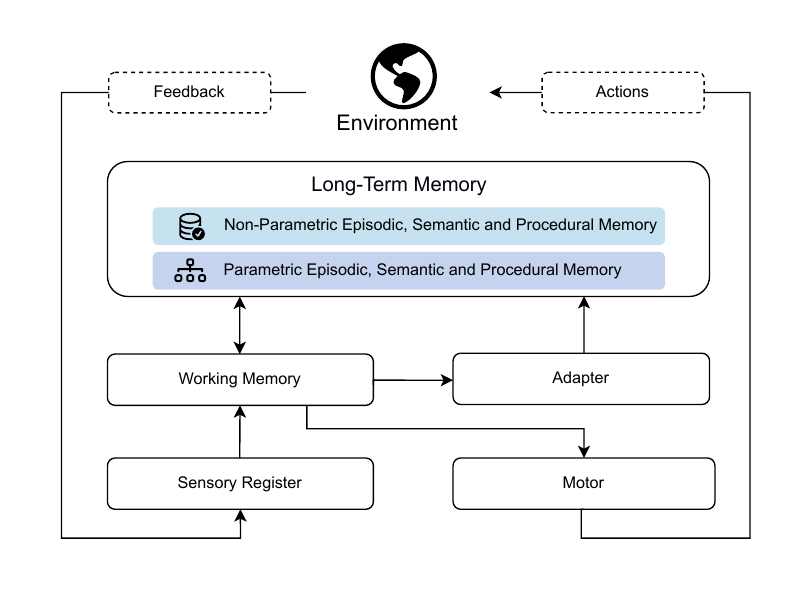}
\caption{Figure of Cognitive Architecture of \textbf{S}elf-\textbf{A}daptive \textbf{L}ong-term \textbf{M}emory (\textbf{SALM}). We propose this framework to integrate theories of AI long-term memory (Sec.~\ref{sec:founddation_ai_memory} and Sec.~\ref{sec:hierarchy_ai_memory}). This framework addresses the limitations of long-term memory modules in current cognitive architectures and has the potential for greater adaptability than the human brain's long-term memory processing mechanisms, positioning it as a guiding framework for the next generation of long-term memory driven AI systems (Sec.~\ref{sec:hierarchy_ai_memory:cognitive_arc}).}
\label{fig:ca_self_adaptive_long_term_memory}
\end{figure}

\begin{figure*}
\centering
\includegraphics[scale=0.5]{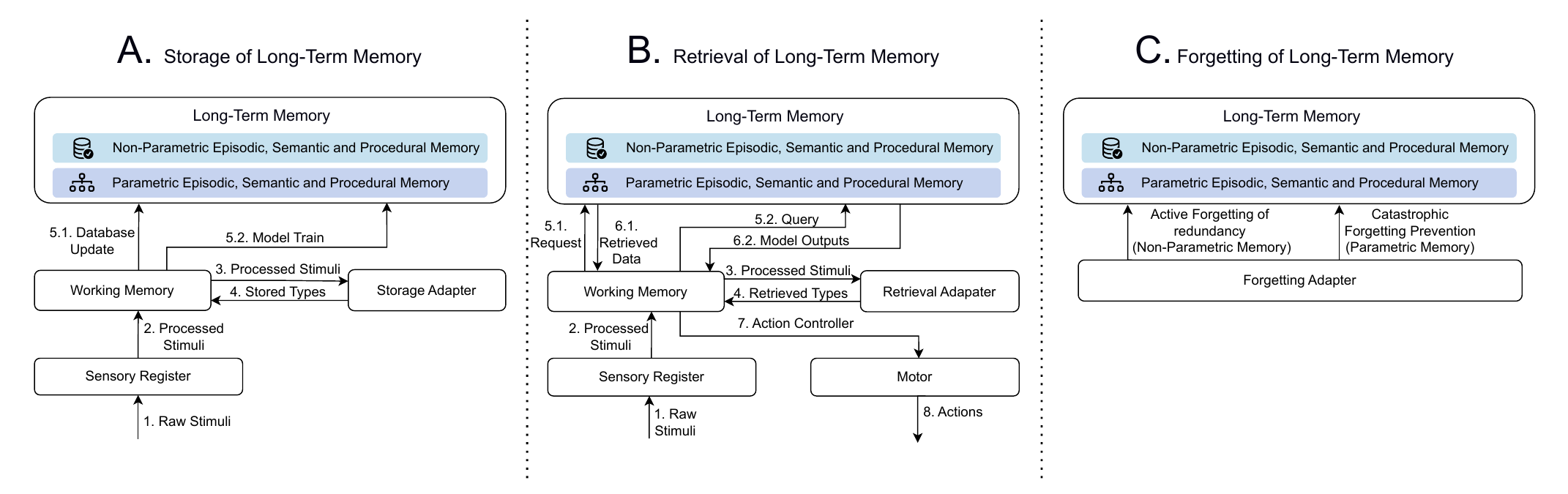}
\caption{Detailed steps of three processing stages of long-term memory in Cognitive Architecture of \textbf{S}elf-\textbf{A}daptive \textbf{L}ong-term \textbf{M}emory (\textbf{SALM}). These stages represent a hypothesized sequence of interactions among components, which may vary in practical implementations.}
\label{fig:process_ca_self_adaptive_long_term_memory}
\end{figure*}

To address these limitations, and building on the findings in this survey, we introduce the Cognitive Architecture of \textbf{S}elf-\textbf{A}daptive \textbf{L}ong-term \textbf{M}emory (\textbf{SALM}) in Fig.~\ref{fig:ca_self_adaptive_long_term_memory}. 
\textbf{SALM} is a theoretical framework introduced to inspire the development of next-generation AI memory systems.
We elaborate on this new concept in the rest of this section.

\textbf{SALM} consists of four key components: a sensory register for capturing stimuli, a working memory for processing them, a long-term memory module for permanent storage, and a motor module for executing actions.
\textbf{SALM} supports all six types of long-term memory (detailed in Sec.~\ref{sec:hierarchy_ai_memory}). 
Inspired by adaptive software~\cite{oreizy1999architecture,salehie2009self} and AI systems~\cite{zhao2022process,zhou2023episodic,wang2023incorporating,mallen2023trust}, we incorporate adapters into SALM to achieve adaptive tuning in storage, retrieval, and forgetting processes. This approach utilizes a feedback loop to optimize the entire memory system with methods such as reinforcement learning~\cite{kaelbling1996reinforcement,liu2021automated,liu2022feature}.
% optimization methods, such as reinforcement learning~\cite{kaelbling1996reinforcement,liu2021automated,liu2022feature}. 
These adapters' functionalities are formulated as follows:

\begin{j_itemize}

    \item[$\bullet$] \textbf{Adaptive Storage.} Part A of Fig.~\ref{fig:process_ca_self_adaptive_long_term_memory} illustrates the steps involved in long-term memory storage within SALM. The \textit{storage adapter} evaluates whether external stimuli should be integrated into long-term memory, filtering out corrupted data, poisoned inputs, and privacy-sensitive information~\cite{gibney2024ai,shumailov2024ai,steinhardt2017certified,cina2023wild,krishnamurthy2011privacy,pan2020privacy,brown2022does}. It also determines the appropriate storage format; for example, it may switch to non-parametric memory when the capacity for parametric storage is reached or when handling time-sensitive and long-tail information~\cite{kaplan2020scaling,zhai2022scaling,kasai2024realtime,mallen2023trust}. The \textit{storage adapter} employs self-adaptive mechanisms, such as monitoring task accuracy post-storage, to refine its filtering and storage decisions. Data that improves accuracy reinforces the learning process, while data that reduces accuracy serves as a negative example.

    \item[$\bullet$] \textbf{Adaptive Retrieval.} The steps for long-term memory retrieval in \textbf{SALM} are outlined in Part B of Fig.~\ref{fig:process_ca_self_adaptive_long_term_memory}. Non-parametric memory retrieval can be achieved through sparse or dense methods~\cite{sparck1972statistical,robertson2004simple,bm252009robertson,codebert2020zhangyin,clip2021radford,yusong2023audio} (Sec.~\ref{sec:AI_memory:non_parametric:memory_retrieval}), while parametric memory relies on forward propagation (Sec.~\ref{sec:AI_memory:parametric:memory_retrieval}). The \textit{retrieval adapter} assesses the necessity of memory retrieval based on external stimuli. If retrieval is needed, it identifies suitable memory forms and guides the retrieval to enhance actions. Additionally, the \textit{retrieval adapter} learns to trigger memory retrieval in context~\cite{asai2023self,wang2023skr,mallen2023trust}, adjusts parameters for various scenarios~\cite{zhou2023episodic}, and selects relevant memory based on effectiveness for the target task~\cite{asai2023self}. It also helps resolve conflicts and hallucinations in retrieved memory~\cite{zhang2023siren,xu2024knowledge} and can enhance query through techniques like LLM-based augmentation~\cite{gao2023precise,pryzant2023automatic,maetal2023query,shinn2024reflexion,liu2024query,wangetal2023query2doc,zheng2023take,shaoetal2023enhancing} and reinforcement learning~\cite{kaelbling1996reinforcement,rennie2017self,mohankumar2021diversity}.

    \item[$\bullet$] \textbf{Adaptive Forgetting.} The long-term memory forgetting process in \textbf{SALM} is depicted in Part C of Fig.~\ref{fig:process_ca_self_adaptive_long_term_memory}. In non-parametric memory, the \textit{forgetting adapter} enhances the effectiveness of active forgetting to improve essential information retrieval (Sec.~\ref{sec:AI_memory:non_parametric:memory_forgetting}). This enhancement can be achieved by refining the selection of forgetting targets based on specific objectives. For parametric memory, \textit{forgetting adapter} primarily prevents catastrophic forgetting~\cite{aleixo2023catastrophic} through various methods. These may include optimizing the selection of rehearsal samples~\cite{lopez2017gradient,nguyen2018variational,de2019episodic,prabhu2020gdumb,roscow2021learning} and identifying high-importance parameter subsets~\cite{Adel2020Continual,xue2022meta}. 

\end{j_itemize}
  
  In summary, \textbf{SALM} has the potential to serve as the theoretical framework for the next generation of AI systems driven by long-term memory. It specifically addresses the limitations of existing cognitive architectures, particularly their long-term memory modules. Unlike the human brain, which relies on evolution rather than learning to develop memory processing mechanisms and is susceptible to memory forgetting due to interference from similar information~\cite{sweller2003evolution,manns2006evolution,allen2013evolution,shiffrin1969memory}, AI's long-term memory can be intentionally designed to be more adaptive using this framework.
  
\section{Next Steps of AI Long-term Memory}
\label{sec:next_steps}

In previous sections, we provided a comprehensive survey of AI long-term memory (Sec.~\ref{sec:founddation_ai_memory} and Sec.~\ref{sec:hierarchy_ai_memory}) and introduced a novel cognitive architecture incorporating an adapter mechanism for the development of long-term memory driven AI systems (Sec.~\ref{sec:hierarchy_ai_memory:cognitive_arc}). 
This section outlines the future directions for AI long-term memory. 
We begin by examining measurement methods for long-term memory processing across various scenarios (Sec.~\ref{sec:next_steps:measures}) and then explore the potential applications where AI long-term memory can be effectively utilized (Sec.~\ref{sec:next_steps:application}).

\subsection{Measures of AI Long-term Memory}
\label{sec:next_steps:measures}

The development of AI long-term memory modules necessitates the measurement of their performance. To evaluate the effectiveness of long-term memory storage, retrieval, and forgetting, it is essential to employ specific metrics. A direct approach involves developing measurement methods that assess the performance of the relevant tasks. 
For example, precision, recall, and the F1 score~\cite{yacouby2020probabilistic} can be used to evaluate the effectiveness of data filtering in parametric memory storage, as described in Sec.~\ref{sec:AI_memory:parametric:storage}. 
This evaluation encompasses the modules' ability to handle challenges such as low-quality data~\cite{gibney2024ai,shumailov2024ai} and poisoned data~\cite{steinhardt2017certified,cina2023wild}. 
Similarly, the NDCG metric~\cite{wang2013theoretical} is suitable for evaluating approximate nearest neighbor search~\cite{arya1993approximate,indyk1998approximate,malkov2018efficient,spann2021chen} as detailed in Sec.~\ref{sec:AI_memory:non_parametric:memory_retrieval}.

However, high performance in a specific task does not necessarily lead to improved outcomes in subsequent downstream tasks. For instance, within the Retrieval-Augmented Generation (RAG) framework~\cite{gao2023retrieval,zhao2024retrieval}, a retriever that demonstrates high ``recall'' in a particular test set may not enhance the performance of subsequent question-answering tasks. Consequently, it is more reliable to evaluate these modules directly based on the objectives of the target task.

On the other hand, in Sec.~\ref{sec:hierarchy_ai_memory:cognitive_arc}, we introduced \textbf{SALM} that includes different types of long-term memory modules and corresponding processing mechanisms.
This framework allows AI system to dynamically adjust its long-term memory processing in response to environmental feedback, thus possessing potential to enhance its adaptability beyond that of the human brain. Nevertheless, qualitative analysis is required to assess the effectiveness of this framework in future studies.
For example, ablation studies~\cite{sheikholeslami2019ablation} can be performed to understand the contribution of each individual long-term memory module and the various adapters employed during long-term memory processing stages.

Furthermore, \textbf{SALM} consists of multiple components. A straightforward approach is to build, train, and then assemble each component into a multi-stage pipeline. However, this method may result in weaker inter-module dependencies compared to an end-to-end integrated system, potentially lowering generalization despite reducing training complexity~\cite{sukhbaatar2015end,glasmachers2017limits}. Therefore, it is crucial to conduct comparative experiments to identify the better-performing implementation.

In summary, to effectively verify the efficacy of long-term memory storage, retrieval, and forgetting strategies, it is crucial to develop target-task-driven metrics. These metrics are indispensable for enhancing the long-term memory management capabilities of the corresponding AI systems. Furthermore, it is necessary to conduct a series of ablation studies and comparative experiments to identify the appropriate modules and implementation approaches within the proposed \textbf{SALM}.

\subsection{Application of AI Long-term Memory}
\label{sec:next_steps:application}

In this section, we discuss potential applications where AI long-term memory can be effectively utilized, focusing on two representative examples. 
First, AI long-term memory is highly effective in multi-modal systems, such as video understanding, where it enhances the system's ability to comprehend video content by incorporating historical user information. 
Second, AI long-term memory holds significant promise in advancing fields like computational neuroscience and social science, where it can be employed to simulate human memory mechanisms.

\begin{j_itemize}

    \item[$\bullet$] \textbf{Video Understanding.} Video understanding refers to tasks related to video analysis such as recognition of entities and actions within videos and video content summarization~\cite{jiao2021new,wu2021towards,apostolidis2021video,ju2022prompting}. 
    This field has broad applications, such as in movie and surveillance analysis, autonomous driving, and aerospace remote sensing systems. Incorporating long-term memory significantly enhances video understanding by enabling the storage and retrieval of key video information for subsequent tasks~\cite{cheng2022xmem,shen2023encode}.
    A typical video understanding pipeline driven by long-term memory converts video frames into feature vectors, which can later be retrieved to support various tasks~\cite{cheng2022xmem,mezghan2022memory,shen2023encode,song2023moviechat,he2024ma}. These frames contain situational information from scenes, which can be regarded as episodic memory. Moreover, some video understanding systems improve their performance by integrating external information sources (semantic memory), such as video clip captions~\cite{yang2023care} and related videos~\cite{jilan2024egoInstructor}.
    Our proposed \textbf{SALM} offers a promising approach for developing video understanding systems capable of managing various types of AI long-term memory. The performance of \textbf{SALM}-based systems can be evaluated using various datasets related to video understanding. For instance, the Ego4D dataset~\cite{grauman2022ego4d} captures daily activities from a first-person perspective, supporting tasks such as indexing past experiences, analyzing present interactions, and anticipating future activities. The Ego-Exo4D dataset~\cite{grauman2024ego}, which includes both egocentric and exocentric views of skilled human activities, is designed for tasks like activity understanding, proficiency estimation, cross-view translation, and 3D hand/body pose estimation. The Replica dataset~\cite{straub2019replica} is commonly used for tasks like 2D/3D semantic segmentation and geometric inference in indoor spaces. Additionally, tools like Project Aria~\cite{engel2023project} contribute to the development of datasets by enabling the recording and streaming of egocentric multi-modal data.

    \item[$\bullet$] \textbf{Human Cognition Simulation.} Simulating human cognition can advance research in computational neuroscience and social science, with long-term memory being one of the most critical components of cognition. Successfully replicating human long-term memory through automated models is highly valued.
    Some computational models and functions can be utilized to simulate the mechanisms of long-term memory. For example, Heald et al.~\cite{heald2023contextual} demonstrated that Bayesian models could be used to compare observed outcomes with known contexts, akin to referencing a form of memory. These models generate posterior probabilities, where lower values may trigger memory storage or updating processes. Savin et al.~\cite{savin2011two} modeled the modules for familiarity and recollection. In their model, the familiarity module uses independent weights to evaluate the age of memory, while the recollection module restores stored memory by integrating these evaluations with a weight matrix and specific cues. Moreover, as mentioned in Sec.~\ref{sec:human_memory:processing:forgetting}, the forgetting process of human memory can be modeled by various types of functions, such as the exponential function~\cite{white1985characteristics} and the power function~\cite{wixted1997genuine}. These examples indicate that the processing mechanisms of human memory can be accurately simulated by some descriptive functions. Some adaptive tuning techniques, such as those proposed by Heydari et al.~\cite{heydari2019softadapt} and Mohanty et al.~\cite{mohanty2021adaptive}, have the potential to refine these descriptive functions. Additionally, modeling human long-term memory enhances the fidelity of sandboxes of society. In sandboxes of society based on LLMs, agents equipped with long-term memory can better simulate human social behaviors, making them valuable subjects for sociological and psychological research~\cite{lei2023recagent,park2023generative,he2024afspp}. These agents can reflect on initial observations to develop new understandings, converting episodic memory into semantic memory. Our proposed \textbf{SALM} can enhance the effectiveness of sandboxes of society by selecting appropriate long-term memory processing mechanisms for different social scenarios.

\end{j_itemize}

In summary, long-term memory of AI is essential for a wide range of applications. Its integration can substantially advance the development of related fields.
Beyond the two applications aforementioned, AI long-term memory can also enhance various domains such as content personalization  (e.g. recommender systems~\cite{zhao2020memory,ko2022survey,wu2023survey}) and motion control (e.g. robotics~\cite{pierson2017deep,zhao2020sim,soori2023artificial}).

\section{Conclusion}

In this paper, we present a narrative review of long-term memory in both human brains and AI systems. We begin by systematically analyzing the taxonomies and reviewing relevant works of human and AI long-term memory. Following this, we establish and summarize the mapping relationships between these two forms of memory. Leveraging these findings, we introduce the Cognitive Architecture of \textbf{S}elf-\textbf{A}daptive \textbf{L}ong-term \textbf{M}emory (\textbf{SALM}), designed to address the limitations of the long-term memory modules in current cognitive architectures. This framework offers potential for greater adaptability than the human brain's long-term memory processing mechanisms, making it a promising framework for the next generation of AI systems driven by long-term memory. Additionally, we explore the importance of target-task-driven metrics in managing AI long-term memory and highlight the critical role AI long-term memory can play in applications such as video understanding and human cognition simulation.

\renewcommand{\thesection}{\Alph{section}} % Ensure sections start with letters A, B, etc.
\renewcommand{\thesubsection}{\thesection.\arabic{subsection}} % Format subsections as A.1, B.1, etc.

\printbibliography

\end{document}